\documentclass[10pt,twocolumn,letterpaper]{article}

\usepackage{cvpr}
\usepackage{times}
\usepackage{epsfig}
\usepackage{graphicx}
\usepackage{amsmath}
\usepackage{amssymb}

\usepackage{arydshln}
\usepackage[dvipsnames,table,xcdraw]{xcolor}
\usepackage{soul}
\usepackage{url}
\usepackage{booktabs}
\usepackage{multirow}
\usepackage{float}
\usepackage{dblfloatfix}
\usepackage{wrapfig}
\usepackage{algorithm}
\usepackage{capt-of}
\usepackage[noend]{algpseudocode}
\newcommand\blfootnote[1]{%
  \begingroup
  \renewcommand\thefootnote{}\footnote{#1}%
  \addtocounter{footnote}{-1}%
  \endgroup
}%

\usepackage[breaklinks=true,bookmarks=false]{hyperref}

\newcommand{\ignore}[1]{}
\newcommand\T{\rule{0pt}{2.6ex}}       
\newcommand\B{\rule[-1.2ex]{0pt}{0pt}} 

\newcommand{\parahead}[1]{\textbf{#1}:\ }

\cvprfinalcopy 

\def\cvprPaperID{1146} 

\begin{document}

\title{On Implicit Filter Level Sparsity in Convolutional Neural Networks}

\author{\vspace{-0.5cm}Dushyant Mehta$^{1,3}$\: Kwang In Kim$^{2}$\: Christian Theobalt$^{1,3}$
\and 
$^1$MPI For Informatics   \: $^2$UNIST  \: $^3$Saarland Informatics Campus \vspace{-0.5cm}}
\maketitle

\begin{abstract}
We investigate \emph{filter level} sparsity that emerges in convolutional neural networks (CNNs) which employ Batch Normalization and ReLU activation, and are trained with adaptive gradient descent techniques and L2 regularization or weight decay. We conduct an extensive experimental study casting our initial findings into hypotheses and conclusions about the mechanisms underlying the emergent filter level sparsity. 
This study allows new insight into the performance gap obeserved between adapative and non-adaptive gradient descent methods in practice. Further, analysis of the effect of training strategies and hyperparameters on the sparsity leads to practical suggestions in designing CNN training strategies enabling us to explore the tradeoffs between feature selectivity, network capacity, and generalization performance. Lastly, we show that the implicit sparsity can be harnessed for neural network speedup at par or better than explicit sparsification / pruning approaches, with no modifications to the typical training pipeline required.

\end{abstract}

\section{Introduction}
In this work we show that filter\footnote{Filter refers to the weights and the nonlinearity associated with a particular feature, acting together as a unit. We use filter and feature interchangeably throughout the document.} level sparsity emerges in certain types of feedforward convolutional neural networks. 

In networks which employ Batch Normalization and ReLU activation, after training, certain filters are observed to not activate for any input. Importantly, the sparsity emerges in the presence of non sparsity inducing regularizers such as L2 and weight decay (WD), and vanishes when regularization is removed. We investigate how this sparsity manifests under different hyperparameter settings, and propose an experimentally backed hypothesis for the cause of this emergent sparsity, and the implications of our findings. 

We find that adaptive flavours of SGD produce a higher degree of sparsity than (m)SGD, both with L2 regularization and weight decay (WD). Further, L2 regularization results in a higher degree of sparsity with adaptive methods than weight decay. Additionally, we show that a multitude of seemingly unrelated factors such as mini-batch size, network size, and task difficulty impact the extent of sparsity. 

These findings are important in light of contemporary attempts to explain the performance gap between (m)SGD and adaptive variants. Any theoretical and practical explorations towards explaining the performance gap between SGD and adaptive variants should account for this inadvertent reduction in network capacity when using adaptive methods, which interplays with both the test accuracy and the generalization gap. 
Contemporaneous work \cite{yaguchi2018adam} has also observed that Adam induces filter sparsity in ReLU networks, but lacks a thorough investigation of the causes.

Through a systematic experimental study, we hypothesize that the emergence of sparsity is the direct result of a disproportionate relative influence of the regularizer (L2 or WD) viz a viz the gradients from the primary training objective of ReLU networks. Multiple factors subtly impact the relative influence of the regularizer in previously known and unknown ways, and various hyperparameters and design choices for training neural networks interplay via these factors to impact the extent of emergent sparsity. 
\blfootnote{This work was funded by the ERC Consolidator Grant 4DRepLy (770784). }


We show that understanding the impact of these design choices yields useful and readily controllable sparsity which can be leveraged for considerable neural network speed up, without trading the generalization performance and without requiring any explicit pruning~\cite{molchanov2017pruning,li2017pruning} or sparsification~\cite{liu2017learning} steps. The implicit sparsification process can remove 70-80\% of the convolutional filters from VGG-16 on CIFAR10/100, far exceeding that for \cite{li2017pruning}, and performs comparable to \cite{liu2017learning} for VGG-11 on ImageNet. 

\section{Observing Filter Sparsity in CNNs}
\label{sec:basic_setup}
We begin with the setup for our initial experiments, and present our primary findings. In subsequent sections we further probe the manifestation of filter sparsity, and present an experimentally backed hypothesis regarding the cause.
\subsection{Setup and Preliminaries}
\begin{figure}
\centering
  \includegraphics[width=0.85\columnwidth]{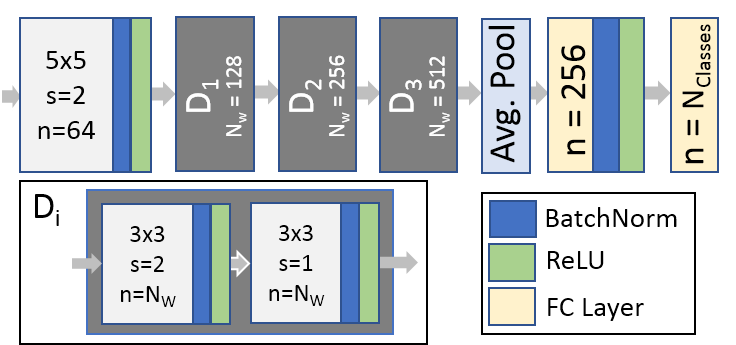}
  \caption{\textbf{BasicNet}: Structure of the basic convolution network studied in this paper. We refer to the individual convolution layers as C1-7. The fully connected head shown here is for CIFAR10/100 and ObjectNet3D~\cite{xiang2016objectnet3d} experiments, and a different fully-connected structure is used for TinyImageNet and ImageNet.}
  \label{fig:basic_net}
  \vspace{-0.5cm}
\end{figure}

Our basic setup is comprised of a 7-layer convolutional network with 2 fully connected layers as shown in Figure~\ref{fig:basic_net}. The network structure is inspired by VGG \cite{simonyan2014very}, but is more compact. We refer to this network as \emph{BasicNet} in the rest of the document. We use a variety of gradient descent approaches, a mini-batch size of 40, with a method specific base learning rate for 250 epochs, and scale down the learning rate by 10 for an additional 75 epochs. We train on CIFAR10 and CIFAR 100 \cite{krizhevsky2009learning}, with normalized images, and random horizontal flips. Xavier initialization \cite{glorot2010understanding} is used for the network weights, with the appropriate gain for ReLU. The base learning rates and other hyperparameters are as follows: Adam (1e-3, $\beta_1$=0.9, $\beta_2$=0.99, $\epsilon$=1e-8), Adadelta (1.0, $\rho$=0.9, $\epsilon$=1e-6), SGD (0.1, momemtum=0.9), Adagrad (1e-2). Pytorch~\cite{paszke2017automatic} is used for training, and we study the effect of varying the amount and type of regularization on the extent of sparsity and test error in Table \ref{tbl:all_optim}.

\parahead{L2 regularization vs. Weight Decay}
We make a distinction between L2 regularization and weight decay. For a parameter $\theta$ and regularization hyperparameter $1>\lambda\geq0$, weight decay multiplies $\theta$ by $(1 - \lambda)$ 
after the update step based on the gradient from the main objective. While for L2 regularization, $\lambda \theta$ is added to the gradient $\nabla L(\theta)$ from the main objective, and the update step is computed using this sum. See \cite{loshchilov2017fixing} for a detailed discussion.

\parahead{Quantifying Feature Sparsity}
We measure the learned feature sparsity in two ways, by per-feature activation and by per-feature scale. For sparsity by activation, for each feature we apply max pooling to the absolute activations over the entire feature plane, and consider the feature inactive if this value does not exceed $10^{-12}$ over the entire \emph{training} corpus. For sparsity by scale, we consider the scale $\gamma$ of the learned affine transform in the Batch Norm \cite{ioffe2015batch} layer. Batch normalization uses additional learned scale $\gamma$ and bias $\beta$ that casts each normalized convolution output $\hat{x}_i$ to $y_i = \gamma \hat{x}_i + \beta$. 
We consider a feature inactive if $|\gamma|$ for the feature is less than $10^{-3}$. Explicitly zeroing the features thus marked inactive does not affect the test error, which ensures the validity of our chosen thresholds. The thresholds chosen are purposefully conservative, and comparable levels of sparsity are observed for a higher feature activation threshold of $10^{-4}$, and a higher $|\gamma|$ threshold of $10^{-2}$.

\subsection{Primary Findings}
\label{sec:primary_findings}
Table~\ref{tbl:all_optim} shows the overall feature sparsity by activation (Act.) and by scale ($\gamma$) for BasicNet. Only convolution features are considered. 
The following are the key observations from the experiments and the questions they raise. These are further discussed in Section~\ref{sec:investigating_sparsity}.

\parahead{1)}
The emergent sparsity relies on the strength of L2 regularization or weight decay. \textbf{No sparsity is observed in the absence of regularization, with sparsity increasing with increasing L2 or weight decay.} \emph{What does this tell us about the cause of sparsification, and how does the sparsity manifest across layers?} 

\parahead{2)}
Regardless of the type of regularizer (L2 or weight decay), \textbf{adaptive methods (Adam, Adagrad, Adadelta) learn sparser representations than SGD for comparable levels of test error}, with Adam showing the most sparsity and most sensitivity to the L2 regularization parameter amongst the ones studied. Adam with L2 sees about 70\% features pruned for CIFAR10, while SGD shows no sparsity for a comparable performance, with a similar trend for CIFAR100, as well as when weight decay is used. \emph{What are the causes for this disparity in sparsity between SGD and adaptive methods?} We will focus on understanding the disparity between SGD and Adam.

\parahead{3)}
SGD has comparable levels of sparsity with L2 regularization and with weight decay (for higher regularization values), while \textbf{for Adam, L2 shows higher sparsity for comparable performance than weight decay} (70\% vs 40\% on CIFAR10, 47\% vs 3\% on CIFAR100). \emph{Why is there a significant difference between the sparsity for Adam with L2 regularization vs weight decay?}

\parahead{4)}
The extent of sparsity decreases on moving from the simple 10 class classification problem of CIFAR10 to the comparatively harder 100 class classification problem of CIFAR100. \emph{What does the task dependence of the extent of sparsity tell us about the origin of the sparsity?}
\renewcommand{\tabcolsep}{1.5pt}
\begin{table}[t]
\centering
\caption{ Convolutional filter sparsity in \emph{BasicNet} trained on CIFAR10/100 for different combinations of regularization and gradient descent methods. Shown are the \% of non-useful / inactive convolution filters, as measured by activation over training corpus (max act. $< 10^{-12}$) and by the learned BatchNorm scale ($|\gamma| < 10^{-03}$), averaged over 3 runs. The lowest test error per optimizer is highlighted, and sparsity (green) or lack of sparsity (red) for the best and near best configurations indicated via text color. L2: L2 regularization, WD: Weight decay (adjusted with the same scaling schedule as the learning rate schedule). Note that for SGD with momentum, L2 and WD are not equivalent~\cite{loshchilov2017fixing}.}
\label{tbl:all_optim}
\resizebox{0.80\linewidth}{!}{
\begin{tabular}{c|c|c|c|c|c|c|c|c|}
\multicolumn{9}{c}{}\\
\cline{2-9}
                              &        & \multicolumn{3}{c}{CIFAR10}               &  &   \multicolumn{3}{c|}{CIFAR100}                      \\ \cline{3-9}
                              &        & \multicolumn{2}{c|}{\% Sparsity}    & Test           &  &   \multicolumn{2}{c|}{\% Sparsity} & Test                     \\
                              & \textbf{L2}       & by Act & by $\gamma$   & Error         &   & by Act & by $\gamma$   & Error              \\ \hline
                               & 2e-03 & 54                           & 54                        & 30.9                                                                                 &  & 69                           & 69                        & 64.8                                                                                 \\
                               & 1e-03 & 27                           & 27                        & 21.8                                                                                  &  & 23                           & 23                        & 47.1                                                                               \\
                               & 5e-04 &  9     &  9  &16.3                                                          &  &                    4            &         4                    &   42.1                                                                                        \\
                               & 2e-04 &  0     &  0  & 13.1                                                          &  &                      0          &    0                         &      38.8                                                                                     \\
                               & 1e-04 &  0     & 0  & 11.8                                                          &  &                {\color[HTML]{9A0000}0}                &           {\color[HTML]{9A0000}0}                  &           \cellcolor[HTML]{9AFF99}37.4                                                                                \\
                               & 1e-05 &  {\color[HTML]{9A0000} 0}     & {\color[HTML]{9A0000} 0}  & \cellcolor[HTML]{9AFF99}10.5                                                          &  &  0     & 0  & 39.0                                                         \\ 
\parbox[t]{2mm}{\multirow{-7}{*}{{\rotatebox[origin=c]{90}{SGD\B}}}}                              & 0        & 0                            & 0                         & 11.3                                                                                  &  & 0                            & 0                         & 40.1                                                                                  \\ \hline
                               & 1e-02 & 82                           & 85                        & 21.3                                                                                &  & 87                           & 85                        & 69.7                                                                               \\
                               & 2e-03 & 88                           & 86                        & 14.7                                                                                 &  & 82                           & 81                        & 42.7                                                                                 \\
                               & 1e-03 & 85                           & 83                        & 13.1                                                                                 &  & {\color[HTML]{009901} 77}    & {\color[HTML]{009901} 76} & 39.0                                                         \\
                               & 1e-04 & {\color[HTML]{009901} 71}    & {\color[HTML]{009901} 70} & \cellcolor[HTML]{9AFF99}10.5                                                          &  & {\color[HTML]{009901} 47}    & {\color[HTML]{009901} 47} & \cellcolor[HTML]{9AFF99}36.6                                                          \\
                               & 1e-05 & {\color[HTML]{009901} 48}    & {\color[HTML]{009901} 48} & 10.7                                                          &  & {\color[HTML]{9A0000} 5}                            & {\color[HTML]{9A0000} 5}                         & 40.6                                                                                  \\
         & 1e-06 &  24    & 24 & 10.9                                                          &  & 0                            & 0                         & 40.5                                                                                 \\ 
 \multirow{-7}{*}{{\rotatebox[origin=c]{90}{Adam~\cite{kingma2014adam}}}}                              & 0        & 3                            & 0                         & 11.0                                                                                  &  & 0                            & 0                         & 40.3                                                                                  \\\hline
                               & 1e-02 & 97                           & 97                        & 36.8                                                                                  &  & 98                           & 98                        & 84.1                                                                                 \\
                               & 2e-03 & 92                           & 92                        & 20.6                                                                                  &  & 89                           & 89                        & 53.2                                                                                  \\
                               & 1e-03 & 89                           & 89                        & 16.7                                                                                  &  & 82                           & 82                        & 46.3                                                                                  \\
                               & 5e-04 &  82     &  82  & 13.6                                                          &  & {\color[HTML]{009901}61}                            & {\color[HTML]{009901}61}                         & 39.1                                                          \\
                               & 2e-04 & {\color[HTML]{009901} 40}     & {\color[HTML]{009901} 40}  & 11.3                                                          &  & {\color[HTML]{9A0000}3}                            & {\color[HTML]{9A0000}3}                         & \cellcolor[HTML]{9AFF99}35.4                                                          \\
\multirow{-6}{*}{{\rotatebox[origin=c]{90}{Adadelta~\cite{zeiler2012adadelta}}}}         & 1e-04 & {\color[HTML]{9A0000} 1}     & {\color[HTML]{9A0000} 1}  & \cellcolor[HTML]{9AFF99}10.2                                                          &  & 1                            & 1                         & 35.9                                                          \\ \hline
                               & 2e-02 & {\color[HTML]{009901} 75}    & {\color[HTML]{009901} 75} & 11.3                                                          &  & 88                           & 88                        & 63.3                                                          \\
                               & 1e-02 & {\color[HTML]{009901} 65}    & {\color[HTML]{009901} 65} & \cellcolor[HTML]{9AFF99}11.2                                                          &  & {\color[HTML]{009901}59}                           & {\color[HTML]{009901}59}                        & 37.2                                                          \\
                               & 5e-03 & {\color[HTML]{009901} 56}    & {\color[HTML]{009901} 56} & 11.3                                                          &  & {\color[HTML]{009901}24}                           & {\color[HTML]{009901}25}                        & \cellcolor[HTML]{9AFF99}35.9                                                          \\
                               & 1e-03 &  27    &  28 & 11.9                                                          &  & 1                           & 1                        & 37.3                                                          \\
\multirow{-5}{*}{{\rotatebox[origin=c]{90}{Adagrad~\cite{duchi2011adaptive}}}}             & 1e-04 & 0                            & 0                         & 13.6                                                                                  &  & 0                            & 0                         & 42.1                                                                                  \\ \hline
                              &        & \multicolumn{3}{c}{CIFAR10}               &  &   \multicolumn{3}{c|}{CIFAR100}                      \\ \cline{3-9}
                              &        & \multicolumn{2}{c|}{\% Sparsity}    & Test           &  &   \multicolumn{2}{c|}{\% Sparsity} & Test                     \\
                              & \textbf{WD}       & by Act & by $\gamma$   & Error         &   & by Act & by $\gamma$   & Error              \\ \hline
                               & 1e-02 & 100                          & 100                       & 90.0                                                                                  &  & 100                          & 100                       & 99.0                                                                                  \\
                               & 1e-03 & 27                           & 27                        & 21.6                                                                                  &  & 23                           & 23                        & 47.6                                                                                  \\
                               & 5e-04 & 8                            & 8                         & 15.8                                                                                  &  & 4                            & 4                         & 41.9                                                                                  \\
                               & 2e-04 & 0    & 0  & 13.3            &  & 0     & 0  & 39.4                                  \\
\multirow{-5}{*}{{\rotatebox[origin=c]{90}{SGD}}}         & 1e-04 & {\color[HTML]{9A0000} 0}     & {\color[HTML]{9A0000} 0}  & \cellcolor[HTML]{9AFF99}12.4                                                          &  & {\color[HTML]{9A0000}0 }                           & {\color[HTML]{9A0000}0}                         & \cellcolor[HTML]{9AFF99}37.7                                                                                  \\ \hline 
                               & 1e-02 & 100                          & 100                       & 82.3                                                                                  &  &       100                         &         100                    &      98.0                                                                                     \\
                               & 1e-03 & 90                           & 90                        & 27.8                                                                                 &  & 81                           & 81                        & 55.3                                                                                 \\
                               & 5e-04 & 81                           & 81                        & 18.1                                                                                  &  &               59                 &      59                       &    43.3                                                                                       \\
                               & 2e-04 & {\color[HTML]{009901} 60}     & {\color[HTML]{009901} 60}  & 13.4                                                          &  & {\color[HTML]{009901} 16}     & {\color[HTML]{009901} 16}  &  37.3                                                          \\
\multirow{-5}{*}{{\rotatebox[origin=c]{90}{Adam~\cite{kingma2014adam}}}}        & 1e-04 & {\color[HTML]{009901} 40}     & {\color[HTML]{009901} 40}  & \cellcolor[HTML]{9AFF99}11.2                                                          &  &    {\color[HTML]{9A0000}3}                            &        {\color[HTML]{9A0000}3}                     &   \cellcolor[HTML]{9AFF99}36.2                                                                                  \\ \hline
\end{tabular}
}
\end{table}
\renewcommand{\tabcolsep}{1.5pt}
\begin{table*}[]
\centering
\caption{Layerwise \% filters pruned from BasicNet trained on CIFAR100, based on the $|\gamma| <10^{-3}$ criteria. Also shown are pre-pruning and post-pruning test error, and the \% of \emph{convolutional} parameters pruned. C1-C7 indicate Convolution layer 1-7, and the numbers in parantheses indicate the total number of features per layer. Average of 3 runs. Color and highlighting indicates high and low sparsity for best and near best test errors, as in Table~\ref{tbl:all_optim}. Refer to the supplementary document for the corresponding table for CIFAR10.}
\label{tbl:layerwise_sparsity}
\resizebox{0.8\linewidth}{!}{
\begin{tabular}{lcccccccccccccc}
                                             \multicolumn{2}{l}{}          & \multicolumn{1}{l}{} & \multicolumn{1}{l}{}           & \multicolumn{1}{l|}{}          & \multicolumn{8}{c|}{\% Sparsity by $\gamma$ or \% Filters Pruned}                                                                                                                                                                & \multicolumn{1}{l|}{\% Param}                                          & \multicolumn{1}{l}{}         \\ \cline{6-13}
                       & \multicolumn{1}{l|}{}         &    Train                  & \multicolumn{1}{c|}{Test}          & \multicolumn{1}{c|}{Test}          & C1                   & C2                   & C3                   & C4                   & C5                   & C6                   & \multicolumn{1}{c|}{C7}    & \multicolumn{1}{c|}{Total}  & \multicolumn{1}{c|}{Pruned}                                            & Pruned                       \\
               & \multicolumn{1}{l|}{}         & Loss           & \multicolumn{1}{c|}{Loss} & \multicolumn{1}{c|}{Err}  & (64)                 & (128)                & (128)                & (256)                & (256)                & (512)                & \multicolumn{1}{c|}{(512)} & \multicolumn{1}{c|}{(1856)} & \multicolumn{1}{c|}{(4649664)}                                           & Test Err.                    \\ \hline
\multicolumn{1}{|l}{}                       & \multicolumn{1}{|c|}{L2: 1e-3} & 1.06                 & \multicolumn{1}{c|}{1.41}      & \multicolumn{1}{c|}{39.0}      & 56                   & 47                   & 43                   & 68                   & 72                   & 91                   & \multicolumn{1}{c|}{85}    & \multicolumn{1}{c|}{76}     & \multicolumn{1}{c|}{95}                        & 39.3 \\
\multicolumn{1}{|l}{}                       & \multicolumn{1}{|c|}{L2: 1e-4} & 0.10                 & \multicolumn{1}{c|}{1.98}      & \multicolumn{1}{c|}{\cellcolor[HTML]{9AFF99}36.6}      & {\color[HTML]{009901}41}                   & {\color[HTML]{009901}20}                   & {\color[HTML]{9A0000}9}                    & {\color[HTML]{009901}33}                   & {\color[HTML]{009901}34}                   & {\color[HTML]{009901}67}                   & \multicolumn{1}{c|}{\color[HTML]{009901}55}    & \multicolumn{1}{c|}{\color[HTML]{009901}47}     & \multicolumn{1}{c|}{\color[HTML]{009901}74}                        & \cellcolor[HTML]{9AFF99}36.6 \\ \cline{2-15} 
\multicolumn{1}{|l}{}  & \multicolumn{1}{|c|}{WD: 2e-4} & 0.34                 & \multicolumn{1}{c|}{1.56}      & \multicolumn{1}{c|}{37.3}      & \color[HTML]{009901}55                    & \color[HTML]{009901}20                    & \color[HTML]{9a0000}3                    & \color[HTML]{9a0000}4                    & \color[HTML]{9a0000}2                    & \color[HTML]{009901}16                    & \multicolumn{1}{c|}{\color[HTML]{009901}26}     & \multicolumn{1}{c|}{\color[HTML]{009901}16}      & \multicolumn{1}{c|}{\color[HTML]{009901}27}                                                 & 37.3                         \\ 
\multicolumn{1}{|c}{\multirow{-4}{*}{\rotatebox[origin=c]{90}{Adam}}} & \multicolumn{1}{|c|}{WD: 1e-4} & 0.08                 & \multicolumn{1}{c|}{1.76}      & \multicolumn{1}{c|}{\cellcolor[HTML]{9AFF99}36.2}      & {\color[HTML]{009901}38}                   & {\color[HTML]{9a0000}4}                    & {\color[HTML]{9a0000}0}                    & {\color[HTML]{9a0000}0}                    & {\color[HTML]{9a0000}0}                    & {\color[HTML]{9a0000}0}                    & \multicolumn{1}{c|}{\color[HTML]{9a0000}5}     & \multicolumn{1}{c|}{\color[HTML]{9A0000}3}      & \multicolumn{1}{c|}{\color[HTML]{9a0000}4}                                                 & {\cellcolor[HTML]{9AFF99}36.2}                        \\ \hline
\multicolumn{1}{|l}{}                       & \multicolumn{1}{|c|}{L2: 1e-3} & 1.49                 & \multicolumn{1}{c|}{1.78}      & \multicolumn{1}{c|}{47.1}      & 82                   & 41                   & 33                   & 29                   & 33                   & 6                    & \multicolumn{1}{c|}{18}    & \multicolumn{1}{c|}{23}     & \multicolumn{1}{c|}{34}                        & 47.1 \\
\multicolumn{1}{|l}{}                       & \multicolumn{1}{|c|}{L2: 5e-4} & 0.89                 & \multicolumn{1}{c|}{1.69}      & \multicolumn{1}{c|}{42.1}      & 64                    & 3                    & 3                    & 3                    & 2                    & 0                    & \multicolumn{1}{c|}{2}     & \multicolumn{1}{c|}{4}      & \multicolumn{1}{c|}{4}                         & 42.1 \\ \cline{2-15} 
\multicolumn{1}{|l}{}                       & \multicolumn{1}{|c|}{WD: 1e-3} & 1.49                 & \multicolumn{1}{c|}{1.79}      & \multicolumn{1}{c|}{47.6}      & 82                   & 43                   & 31                   & 28                   & 33                   & 6                   & \multicolumn{1}{c|}{17}    & \multicolumn{1}{c|}{23}     & \multicolumn{1}{c|}{34}                                                & 47.6                         \\
\multicolumn{1}{|c}{\multirow{-4}{*}{\rotatebox[origin=c]{90}{SGD}}}  & \multicolumn{1}{|c|}{WD: 5e-4} & 0.89                 & \multicolumn{1}{c|}{1.69}      & \multicolumn{1}{c|}{41.9}      & 66                    & 2                    & 1                   & 4                    & 2                    & 0                    & \multicolumn{1}{c|}{1}     & \multicolumn{1}{c|}{4}      & \multicolumn{1}{c|}{4}                         & 41.9 \\ \hline
\end{tabular}
}
\vspace{-0.2cm}
\end{table*}
\section{A Detailed Look at the Emergent Sparsity}
\label{sec:investigating_sparsity}

\parahead{Possible Cause of Sparsity}
The analysis of Table~\ref{tbl:all_optim} in the preceding section shows that the regularizer (L2 or weight decay) is very likely the cause of the sparsity, with differences in the level of sparsity attributable to the particular interaction of L2 regularizer (and lack of interaction of weight decay) with the update mechanism. The differences between adaptive gradient methods (Adam) and SGD can additionally likely be attributed to differences in the nature of the learned representations between the two. That would explain the higher sparsity seen for Adam in the case of weight decay. 

\parahead{Layer-wise Sparsity}
To explore the role of the regularizer in the sparsification process, 
we start with a layer-wise breakdown of sparsity.
For each of Adam and SGD, we consider both L2 regularization and weight decay in Table~\ref{tbl:layerwise_sparsity} for CIFAR100. The table shows sparsity by scale ($|\gamma| < 10^{-3}$) for each convolution layer. For both optimizer-regularizer pairings we pick the configurations from Table~\ref{tbl:all_optim} with the lowest test errors that also produce sparse features.
For SGD, the extent of sparsity is higher for earlier layers, and decreases for later layers. The trend holds for both L2 and weight decay, from C1-C6. Note that the higher sparsity seen for C7 might be due to its interaction with the fully connected layers that follow. Sparsity for Adam shows a similar decreasing trend from early to middle layers, and increasing sparsity from middle to later layers.

\begin{figure*}
  \centering
  \includegraphics[width=0.90\linewidth]{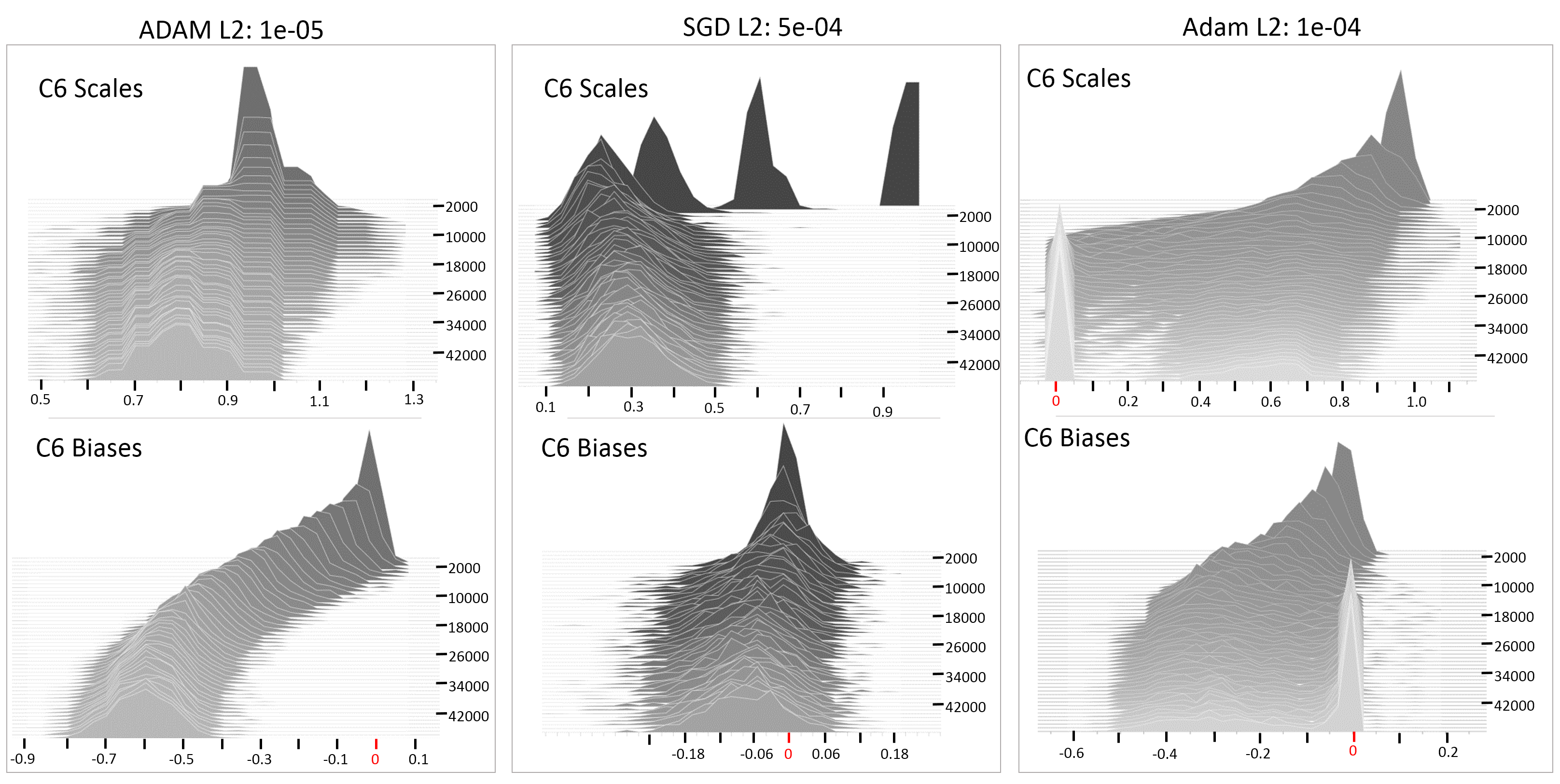}
  \caption{\textbf{Emergence of Feature Selectivity with Adam} The evolution of the learned scales ($\gamma$, top row) and biases ($\beta$, bottom row) 
for layer C6 of \emph{BasicNet} for Adam and SGD as training progresses. Adam has distinctly negative biases, while SGD sees both positive and negative biases. For positive scale values, as seen for both Adam and SGD, this translates to greater feature selectivity in the case of Adam, which translates to a higher degree of sparsification when stronger regularization is used. Note the similarity of the final scale distribution for Adam L2:1e-4 to the scale distributions shown in Figure 4 in \cite{liu2017learning}}
  \label{fig:bias_scale_distribution}
  \vspace{-0.5cm}
\end{figure*}
\parahead{Surprising Similarities to Explicit Feature Sparsification}
In the case of Adam, the trend of layerwise sparsity exhibited is similar to that seen in explicit feature sparsification approaches (See Table 8 in \cite{liu2018rethinking} for Network Slimming \cite{liu2017learning}). 
If we explicitly prune out features meeting the $|\gamma| < 10^{-3}$ sparsity criteria, we still see a relatively high performance on the test set even with 90\% of the convolutional parameters pruned. 
Network Slimming \cite{liu2017learning} uses explicit sparsity constraints on BatchNorm scales ($\gamma$). The similarity in the trend of Adam's emergent layer-wise sparsity to that of explicit scale sparsification motivates us to examine the distribution of the learned scales ($\gamma$) and biases ($\beta$) of the BatchNorm layer in our network. We consider layer C6, and in Figure~\ref{fig:bias_scale_distribution} show the evolution of the distribution of the learned bias and scales as training progresses on CIFAR100. We consider a low L2 regularization value of 1e-5 and a higher L2 regularization value of 1e-4 for Adam, and also show the same for SGD with L2 regularization of 5e-4. The lower regularization values, which do not induce sparsity, would help shed light at the underlying processes without interference from the sparsification process.

\begin{figure*}
\centering
  \includegraphics[width=0.85\linewidth]{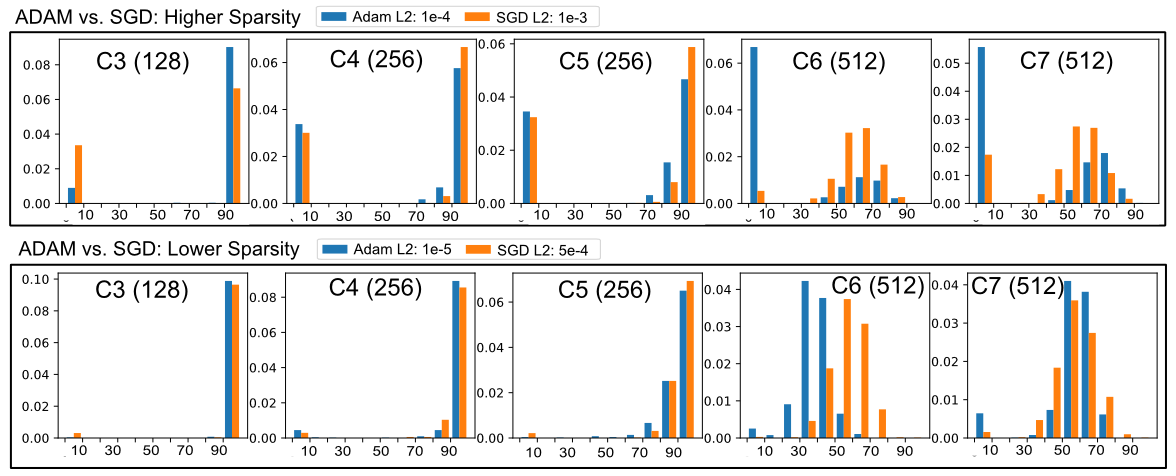}
  \caption{\textbf{Layer-wise Feature Selectivity} Feature universality for CIFAR 100, with Adam and SGD. X-axis shows the universality and Y-axis ($\times 10$) shows the fraction of features with that level of universality. For later layers, Adam tends to learn less universal features than SGD, which get pruned by the regularizer. Please be mindful of the differences in Y-axis scales between plots. Refer to the supplementary document for a similar analysis for CIFAR10}
  \label{fig:cifar_100_feat_selectivity}
  \vspace{-0.5cm}
\end{figure*}
\parahead{Feature Selectivity Hypothesis}
From Figure~\ref{fig:bias_scale_distribution} the differences between the nature of features learned by Adam and SGD become clearer. 
For zero mean, unit variance BatchNorm outputs $\{\hat{x}_i\}_{i=1}^N$ of a particular convolutional kernel, where $N$ is the size of the training corpus, due to the use of ReLU, a gradient is only seen for those datapoints for which $\hat{x}_i > -\beta/\gamma$. Both SGD and Adam (L2: 1e-5) learn positive $\gamma$s for layer C6, however $\beta$s are negative for Adam, while for SGD some of the biases are positive. This implies that all features learned for Adam (L2: 1e-5) in this layer activate for $\leq$ half the activations from the training corpus, while SGD has a significant number of features activate for more than half of the training corpus, i.e., Adam learns more selective features in this layer.
Features which activate only for a small subset of the training corpus, and consequently see gradient updates from the main objective less frequently, continue to be acted upon by the regularizer. If the regularization is strong enough (Adam with L2: 1e-4 in Fig.~\ref{fig:bias_scale_distribution}), or the gradient updates infrequent enough (feature too selective), the feature may be pruned away entirely. 
The propensity of later layers to learn more selective features with Adam would explain the higher degree of sparsity seen for later layers as compared to SGD. Understanding the reasons for emergence of higher feature selectivity in Adam than SGD, and verifying if other adaptive gradient descent flavours also exhibit higher feature selectivity remains open for future investigation. 

\parahead{Quantifying Feature Selectivity}
Similar to feature sparsity by activation, we apply max pooling to a feature's absolute activations over the entire feature plane. For a particular feature, we consider these pooled activations over the entire training corpus and normalize them by the max of the pooled activations over the entire training corpus. We then consider the percentage of the training corpus for which this normalized pooled value exceeds a threshold of $10^{-3}$. We refer to this percentage as the feature's \emph{universality}. A feature's selectivity is then defined as 100-\emph{universality}. Unlike the selectivity metrics employed in literature~\cite{morcos2018importance}, ours is class agnostic. In Figure ~\ref{fig:cifar_100_feat_selectivity}, we compare the `universality' of features learned with Adam and SGD per layer on CIFAR100, for both low and higher regularization values. For the low regularization case, we see that in C6 and C7 both Adam and SGD learn selective features, with Adam showing visibly `more selectivity for C6 (blue bars shifted left). The disproportionately stronger regularization effect of L2 coupled with Adam becomes clearer when moving to a higher regularization value. The selectivity for SGD in C6 remains mostly unaffected, while Adam sees a large fraction (64\%) of the features inactivated (0\% universality). Similarly for C7, the selectivity pattern remains the same on moving from lower regularization to higher regularization, but Adam sees more severe feature inactivation.

\begin{figure*}
  \centering
  \includegraphics[width=0.85\linewidth, trim={0 11cm 13.5cm 0},clip]{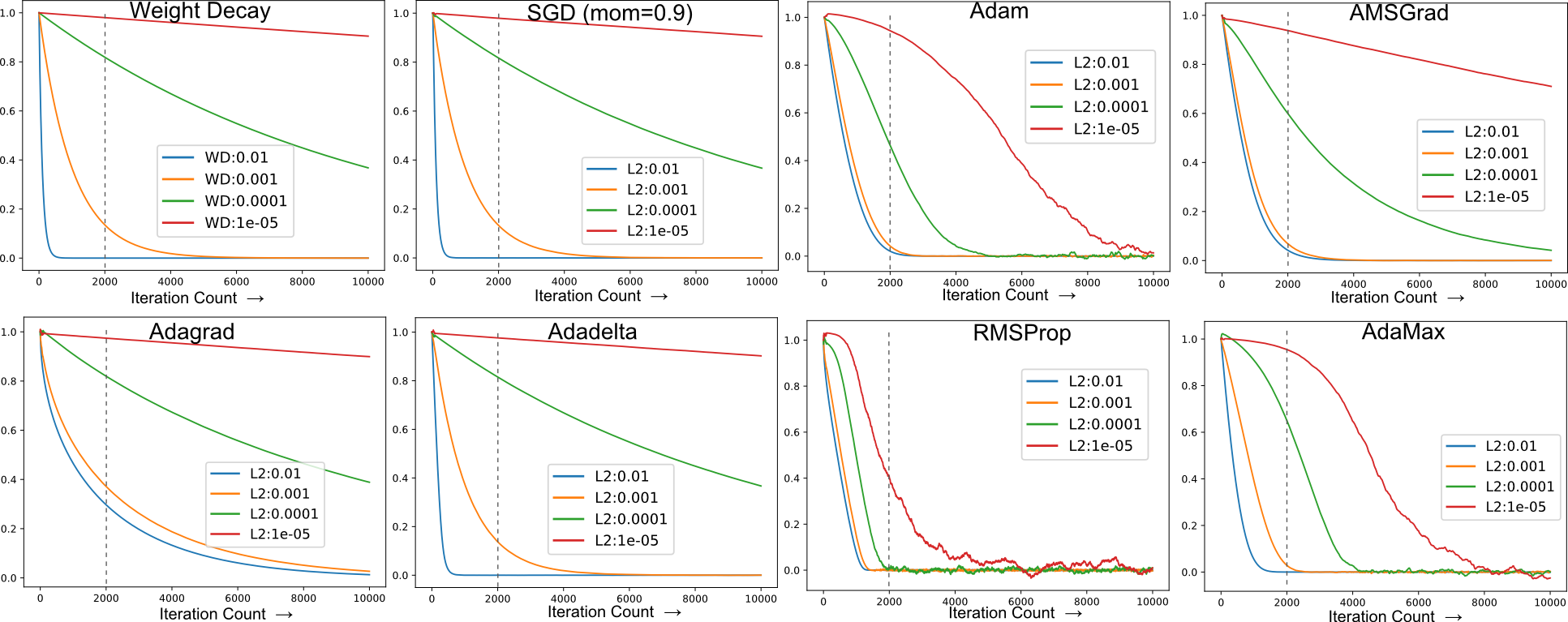}
  \caption{The action of regularization on a scalar value, for a range of regularization values in the presence of simulated low gradients drawn from a mean=0, std=$10^{-5}$ normal distribution. The gradients for the first 100 iterations are drawn from a mean=0, std=$10^{-3}$ normal distribution to emulate a transition into low gradient regime rather than directly starting in a low gradient regime. The learning rate for SGD(momentum=0.9) is 0.1, and the learning rate for ADAM is 1e-3. We show similar plots for other adaptive gradient descent approaches in the supplementary document.
}
  \label{fig:wd_l2_decay}
  \vspace{-0.5cm}
\end{figure*}

\parahead{Interaction of L2 Regularizer with Adam}
Next, we consider the role of the L2 regularizer vs. weight decay.
We study the behaviour of L2 regularization in the low gradient regime for different optimizers. 
Figure~\ref{fig:wd_l2_decay} shows that coupling of L2 regularization with ADAM update equation yields a faster decay than weight decay, or L2 regularization with SGD, even for smaller regularizer values.
This is an additional source of regularization disparity between parameters which see frequent updates and those which don't see frequent updates or see lower magnitude gradients. It manifests for certain adaptive gradient descent approaches.

\parahead{Task `Difficulty' Dependence} As per the hypothesis developed thus far, as the task becomes more difficult, for a given network capacity, we expect the fraction of features pruned to decrease corresponding to a decrease in selectivity of the learned features~\cite{zhou2018revisiting}. This is indeed observed in Table~\ref{tbl:all_optim} for \emph{BasicNet} for all gradient descent methods on moving from CIFAR10 to CIFAR100. For Adam with L2 regularization, 70\% sparsity on CIFAR10 decreases to 47\% on CIFAR 100, and completely vanishes on ImageNet (See Table~\ref{tbl:imagenet_basic}). 
A similar trend is evident for VGG-16 in Tables~\ref{tbl:vgg_cifar} and ~\ref{tbl:vgg16_tinyimagenet}.
In Figure \ref{fig:adam_batch_feat} note the distinct shift towards less selective features in \emph{BasicNet} with increasing task difficulty. 

Since the task difficulty cannot be cleanly decoupled from the number of classes, we devise a synthetic experiment based on grayscale renderings of 30 object classes from ObjectNet3D \cite{xiang2016objectnet3d}. We construct 2 identical sets of $\approx50k$ 64$\times$64 pixel renderings, one with a clean background (BG) and the other with a cluttered BG. We train \emph{BasicNet} with a mini-batch size of 40, and see that as expected there is a much higher sparsity (70\%) with the clean BG set than with the more difficult cluttered set (57\%). See the supplemental document for representative images and a list of the object classes selected. 
\section{Related Work}
\parahead{Effect of L2 regularization vs. Weight Decay for Adam}
Prior work \cite{loshchilov2017fixing} has indicated that Adam with L2 regularization leads to parameters with frequent and/or large magnitude gradients from the main objective being regularized less than the ones which see infrequent and/or small magnitude gradients. Though weight decay is proposed as a supposed fix, we show that there are rather two different aspects to consider. The first is the disparity in effective regularization due to the frequency of updates. 
Parameters which update less frequently would see more regularization steps per actual update than those which are updated more frequently. This disparity would persist even with weight decay due to Adam's propensity for learning more selective features, as detailed in the preceding section. The second aspect is the additional disparity in regularization for features which see low/infrequent gradient, due to the coupling of L2 regularization with Adam.

\parahead{Attributes of Generalizable Neural Network Features}
Dinh et al. \cite{Dinh2017SharpMC} show that the geometry of minima is not invariant to reparameterization, and thus the flatness of the minima may not be indicative of generalization performance \cite{keskar2016large}, or may require other metrics which are invariant to reparameterization.
Morcos et al. \cite{morcos2018importance} suggest based on extensive experimental evaluation that good generalization ability is linked to reduced selectivity of learned features. They further suggest that individual selective units do not play a strong role in the overall performance on the task as compared to the less selective ones. They connect the ablation of selective features to the heuristics employed in neural network feature pruning literature which prune features whose removal does not impact the overall accuracy significantly \cite{molchanov2017pruning,li2017pruning}.  
The findings of Zhou et al. \cite{zhou2018revisiting} concur regarding the link between emergence of feature selectivity and poor generalization performance. They further show that ablation of class specific features does not influence the overall accuracy significantly, however the specific class may suffer significantly. 
We show that the emergence of selective features in Adam, and the increased propensity for pruning the said selective features when using L2 regularization presents a direct tradeoff between generalization performance and network capacity which practitioners using Adam must be aware of.

\parahead{Observations on Adaptive Gradient Descent}
Several works have noted the poorer generalization performance of adaptive gradient descent approaches over SGD. Keskar et al. \cite{keskar2017improving} propose to leverage the faster initial convergence of ADAM and the better generalization performance of SGD, by switching from ADAM to SGD while training. Reddi et al. \cite{reddi2018convergence} point out that exponential moving average of past squared gradients, which is used for all adaptive gradient approaches, is problematic for convergence, particularly with features which see infrequent updates. This short term memory is likely the cause of accelerated pruning of selective features seen for Adam in Figure~\ref{fig:wd_l2_decay}(and other adaptive gradient approaches), and the extent of sparsity observed would be expected to go down with AMSGrad which tracks the long term history of squared gradients. 

\parahead{Feature Pruning/Sparsification}
Among the various explicit filter level sparsification heuristics and approaches ~\cite{li2017pruning,srinivas2015data,hu2016network,theis2018faster,molchanov2017pruning,mozer1989skeletonization,liu2017learning,ye2018rethinking}, some \cite{ye2018rethinking,liu2017learning} make use of the learned scale parameter $\gamma$ in Batch Norm for enforcing sparsity on the filters. 
Ye et al. \cite{ye2018rethinking} argue that BatchNorm makes feature importance less susceptible to scaling reparameterization, and the learned scale parameters ($\gamma$) can be used as indicators of feature importance.
We find that Adam with L2 regularization, owing to its implicit pruning of features based on feature selectivity, makes it an attractive alternative to explicit sparsification/pruning approaches. The link between ablation of selective features and explicit feature pruning is also established in prior work~\cite{morcos2018importance,zhou2018revisiting}.

\section{Further Experiments}
We conduct additional experiments on various datasets and network architectures to show that the intuition developed in the preceding sections generalizes.
Further, we provide additional support by analysing the effect of various hyperparameters on the extent of sparsity. We also compare the emergent sparsity for different networks on various datasets to that of explicit sparsification approaches.

\parahead{Datasets} In addition to CIFAR10 and CIFAR100, we also consider TinyImageNet~\cite{TinyImagenet} which is a 200 class subset of ImageNet~\cite{imagenet_cvpr09} with images resized to 64$\times$64 pixels. The same training augmentation scheme is used for TinyImageNet as for CIFAR10/100. We also conduct extensive experiments on ImageNet. The images are resized to 256$\times$256 pixels. and random crops of size 224$\times$224 pixels used while training, combined with random horizontal flips. For testing, no augmentation is used, and 1-crop evaluation protocol is followed.

\parahead{Network Architectures} 
The convolution structure for \emph{BasicNet} stays the same across tasks, while the fully-connected (\emph{fc}) structure  changes across task. We will use `[$n$]' to indicate an \emph{fc} layer with $n$ nodes. Batch Norm and ReLU are used in between \emph{fc} layers. For CIFAR10/100 we use Global Average Pooling (GAP) after the last convolution layer and the \emph{fc} structure is [256][10]/[256][100], as shown in Figure~\ref{fig:basic_net}. 
For TinyImagenet we again use GAP followed by [512][256][200]. On ImageNet we use average pooling with a kernel size of 5 and a stride of 4, followed by [4096][2048][1000].
For VGG-11/16, on CIFAR10/100 we use [512][10]/[512][100]. For TinyImageNet we use [512][256][200], and for ImageNet we use the structure in~\cite{simonyan2014very}.
For VGG-19, on CIFAR10/100, we use an \emph{fc} structure identical to ~\cite{liu2017learning}.
Unless explicitly stated, we will be using Adam with L2 regularization of 1e-4, and a batch size of 40. When comparing different batch sizes, we ensure the same number of training iterations.

\begin{figure}[ht]
  \includegraphics[width=\linewidth]{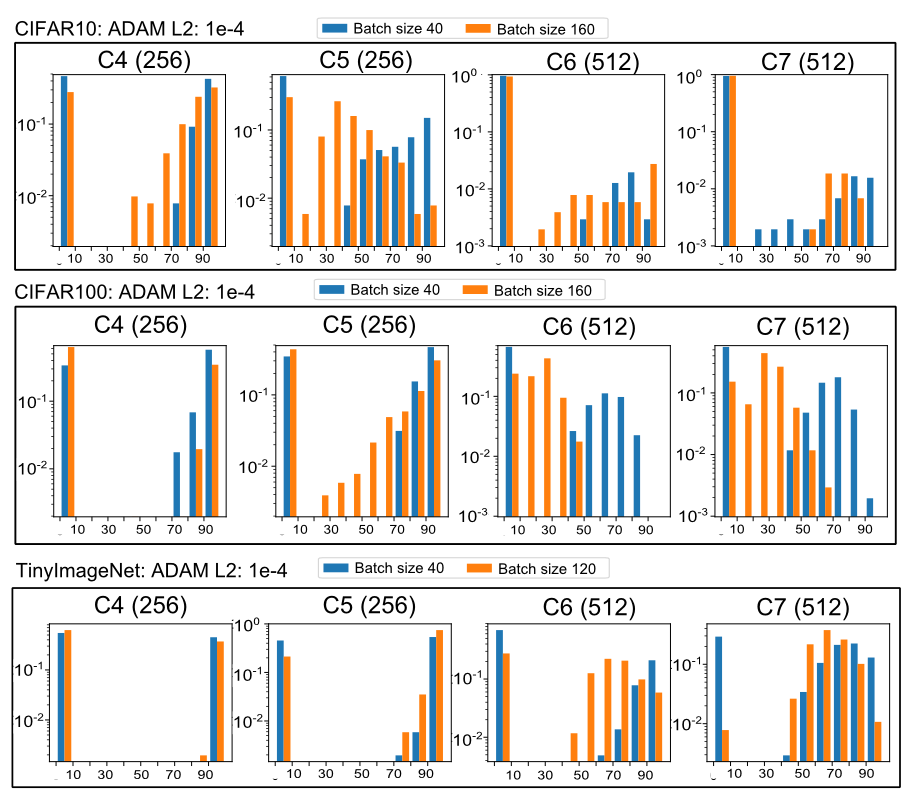}
  \caption{\textbf{Feature Selectivity For Different Mini-Batch Sizes for Different Datasets} Feature universality (1 - selectivity) plotted for layers C4-C7 of \emph{BasicNet} for CIFAR10, CIFAR100 and TinyImagenet. Batch sizes of 40/160 considered for CIFAR, and 40/120 for TinyImagenet.}
  \label{fig:adam_batch_feat}
  \vspace{-0.3cm}
\end{figure}
\subsection{Analysis of Hyperparameters}
Having established in Section~\ref{sec:investigating_sparsity} (Figures~\ref{fig:cifar_100_feat_selectivity} and ~\ref{fig:bias_scale_distribution}) that with Adam, the emergence of sparsity is correlated with feature selectivity, we investigate the impact of various hyperparameters on the emergent sparsity.

\parahead{Effect of Mini-Batch Size} 
Figure \ref{fig:adam_batch_feat} shows the extent of feature selectivity for C4-C7 of \emph{BasicNet} on CIFAR and TinyImageNet for different mini-batch sizes. For each dataset, note the apparant increase in selective features with increasing batch size. However, a larger mini-batch size is not promoting feature selectivity, and rather preventing the selective features from being pruned away by providing more frequent updates. 
This makes the mini-batch size a key knob to control tradeoffs between network capacity (how many features get pruned, which affects the speed and performance) and generalization ability (how many selective features are kept, which can be used to control overfitting).
We see across datasets and networks that increasing the mini-batch size leads to a decrease in sparsity (Tables~\ref{tbl:bat_size},~\ref{tbl:tnyimngnet_basic},~\ref{tbl:imagenet_basic},~\ref{tbl:vgg_cifar},~\ref{tbl:vgg16_tinyimagenet},~\ref{tbl:vgg11_imagenet},~\ref{tbl:vgg19_cifar}). 


\parahead{Network Capacity} 
Task `difficulty' is relative to the network's learning capacity. In the preceding section we directly manipulated the task difficulty, and here we consider variations of \emph{BasicNet} in Table~\ref{tbl:net_width} to study the complementary effect of network capacity. We indicate the architecture presented in Figure~\ref{fig:basic_net} as `64-1x', and consider two variants: `64-0.5x' which has 64 features in the first convolution layer, and half the features of \emph{BasicNet} in the remaining convolution layers, and `32-0.25x' with 32 features in the first channel and a quarter of the features in the remaining layers. The fc-head remains unchanged. We see a consistent decrease in the extent of sparsity with decreasing network width in Table~\ref{tbl:net_width}. Additionally note the decrease in sparsity in moving from CIFAR10 to CIFAR100.
\begin{table}[htb!]
\renewcommand{\tabcolsep}{1.5pt}
\centering
\caption{BasicNet sparsity variation on CIFAR10/100 trained with Adam and L2 regularization.}
\label{tbl:bat_size}
\begin{tabular}{cc|c|c|c|c|c|cccc}
\multicolumn{1}{l}{}                             & \multicolumn{1}{l|}{} & \multicolumn{4}{c|}{CIFAR 10}                                                  & \multicolumn{1}{l|}{} & \multicolumn{4}{c}{CIFAR 100}                                                                                                                 \\
\multicolumn{1}{l|}{}                            & Batch                 & Train                       & Test                        & Test & \%Spar.  &                       & \multicolumn{1}{c|}{Train}                       & \multicolumn{1}{c|}{Test}                        & \multicolumn{1}{c|}{Test} & \%Spar.  \\
\multicolumn{1}{c|}{}                            & Size                  & Loss                        & Loss                        & Err  & by $\gamma$ &                       & \multicolumn{1}{c|}{Loss}                        & \multicolumn{1}{c|}{Loss}                        & \multicolumn{1}{c|}{Err}  & by $\gamma$ \\ \hline
\multicolumn{1}{c|}{}                           & 20                    & {\color[HTML]{9B9B9B} 0.43} & {\color[HTML]{9B9B9B} 0.45} & 15.2 & 82          &                       & \multicolumn{1}{c|}{{\color[HTML]{9B9B9B} 1.62}} & \multicolumn{1}{c|}{{\color[HTML]{9B9B9B} 1.63}} & \multicolumn{1}{c|}{45.3} & 79          \\
\multicolumn{1}{c|}{}                           & 40                    & {\color[HTML]{9B9B9B} 0.29} & {\color[HTML]{9B9B9B} 0.41} & 13.1 & 83          &                       & \multicolumn{1}{c|}{{\color[HTML]{9B9B9B} 1.06}} & \multicolumn{1}{c|}{{\color[HTML]{9B9B9B} 1.41}} & \multicolumn{1}{c|}{39.0} & 76          \\
\multicolumn{1}{c|}{\multirow{-3}{*}{\rotatebox[origin=c]{90}{L2: 1e-3\B}}} & 80                    & {\color[HTML]{9B9B9B} 0.18} & {\color[HTML]{9B9B9B} 0.40} & 12.2 & 80          &                       & \multicolumn{1}{c|}{{\color[HTML]{9B9B9B} 0.53}} & \multicolumn{1}{c|}{{\color[HTML]{9B9B9B} 1.48}} & \multicolumn{1}{c|}{37.1} & 67          \\ \hline
\multicolumn{1}{c|}{}                           & 20                    & {\color[HTML]{9B9B9B} 0.17} & {\color[HTML]{9B9B9B} 0.36} & 11.1 & 70          &                       & \multicolumn{1}{c|}{{\color[HTML]{9B9B9B} 0.69}} & \multicolumn{1}{c|}{{\color[HTML]{9B9B9B} 1.39}} & \multicolumn{1}{c|}{35.2} & 57          \\
\multicolumn{1}{c|}{}                           & 40                    & {\color[HTML]{9B9B9B} 0.06} & {\color[HTML]{9B9B9B} 0.43} & 10.5 & 70          &                       & \multicolumn{1}{c|}{{\color[HTML]{9B9B9B} 0.10}} & \multicolumn{1}{c|}{{\color[HTML]{9B9B9B} 1.98}} & \multicolumn{1}{c|}{36.6} & 46          \\
\multicolumn{1}{c|}{}                           & 80                    & {\color[HTML]{9B9B9B} 0.02} & {\color[HTML]{9B9B9B} 0.50} & 10.1 & 66          &                       & \multicolumn{1}{c|}{{\color[HTML]{9B9B9B} 0.02}} & \multicolumn{1}{c|}{{\color[HTML]{9B9B9B} 2.21}} & \multicolumn{1}{c|}{41.1} & 35          \\
\multicolumn{1}{c|}{\multirow{-4}{*}{\rotatebox[origin=c]{90}{L2: 1e-4\B}}} & 160                   & {\color[HTML]{9B9B9B} 0.01} & {\color[HTML]{9B9B9B} 0.55} & 10.6 & 61          &                       & \multicolumn{1}{c|}{{\color[HTML]{9B9B9B} 0.01}} & \multicolumn{1}{c|}{{\color[HTML]{9B9B9B} 2.32}} & \multicolumn{1}{c|}{44.3} & 29          \\ \hline
\end{tabular}
\label{tbl:bat_size}
\vspace{-0.4cm}
\end{table}
\begin{table}[ht!]
\renewcommand{\tabcolsep}{1.5pt}
\centering
\caption{Convolutional filter sparsity for BasicNet trained on TinyImageNet, with different mini-batch sizes. 
}
\label{tbl:tinyimagenet_basic}
\begin{tabular}{c|c|c|c|c|c|c}
                       & Batch & Train                       & Val                         & Top 1    & Top 5    & \% Spar. \\
                       & Size  & Loss                        & Loss                        & Val Err. & Val Err. & by $\gamma$ \\ \hline
SGD                    & 40    & {\color[HTML]{9B9B9B} 0.02} & {\color[HTML]{9B9B9B} 2.63} & 45.0     & 22.7     & 0           \\ \hline
                       & 20    & {\color[HTML]{9B9B9B} 1.05} & {\color[HTML]{9B9B9B} 2.13} & 47.7     & 22.8     & 63          \\
                       & 40    & {\color[HTML]{9B9B9B} 0.16} & {\color[HTML]{9B9B9B} 2.96} & 48.4     & 24.7     & 48          \\
\multirow{-3}{*}{Adam} & 120   & {\color[HTML]{9B9B9B} 0.01} & {\color[HTML]{9B9B9B} 2.48} & 48.8     & 27.4     & 26          \\ \hline
\end{tabular}
\label{tbl:tnyimngnet_basic}
\vspace{-0.4cm}
\end{table}
\begin{table}[ht!]
\renewcommand{\tabcolsep}{1.5pt}
\centering
\caption{Convolutional filter sparsity of BasicNet on ImageNet.}
\label{tbl:imagenet_basic}
\resizebox{0.80\linewidth}{!}{
\begin{tabular}{c|c|c|c|c|c|c}
                       & Batch & Train                       & Val                         & Top 1    & Top 5    & \% Sparsity \\
                       & Size  & Loss                        & Loss                        & Val Err. & Val Err. & by $\gamma$ \\ \hline
                       & 64    & {\color[HTML]{9B9B9B} 2.05} & {\color[HTML]{9B9B9B} 1.58} & 38.0     & 15.9     & 0.2          \\
                       & 256    & {\color[HTML]{9B9B9B} 1.63} & {\color[HTML]{9B9B9B} 1.35} & 32.9     & 12.5     & 0.0          \\
\end{tabular}
}
\label{tbl:imagenet_basic}
\vspace{-0.4cm}
\end{table}
\begin{table}[ht!]
\renewcommand{\tabcolsep}{1.5pt}
\centering
\caption{Effect of varying the number of features in BasicNet. }
\begin{tabular}{c|cccc|c|cccc}
         & \multicolumn{4}{c}{CIFAR 10}                                                                                                                  &  & \multicolumn{4}{c}{CIFAR 100}                                                                                                                 \\ \cline{2-10} 
Net      & \multicolumn{1}{c|}{Train}                       & \multicolumn{1}{c|}{Test}                        & \multicolumn{1}{c|}{Test} & \%Spar.  &  & \multicolumn{1}{c|}{Train}                       & \multicolumn{1}{c|}{Test}                        & \multicolumn{1}{c|}{Test} & \%Spar.  \\
Cfg.     & \multicolumn{1}{c|}{Loss}                        & \multicolumn{1}{c|}{Loss}                        & \multicolumn{1}{c|}{Err}  & by $\gamma$ &  & \multicolumn{1}{c|}{Loss}                        & \multicolumn{1}{c|}{Loss}                        & \multicolumn{1}{c|}{Err}  & by $\gamma$ \\ \hline
64-1x    & \multicolumn{1}{c|}{{\color[HTML]{9B9B9B} 0.06}} & \multicolumn{1}{c|}{{\color[HTML]{9B9B9B} 0.43}} & \multicolumn{1}{c|}{10.5} & 70          &  & \multicolumn{1}{c|}{{\color[HTML]{9B9B9B} 0.10}} & \multicolumn{1}{c|}{{\color[HTML]{9B9B9B} 1.98}} & \multicolumn{1}{c|}{36.6} & 46          \\
64-0.5x  & \multicolumn{1}{c|}{{\color[HTML]{9B9B9B} 0.10}} & \multicolumn{1}{c|}{{\color[HTML]{9B9B9B} 0.41}} & \multicolumn{1}{c|}{11.0} & 51          &  & \multicolumn{1}{c|}{{\color[HTML]{9B9B9B} 0.11}} & \multicolumn{1}{c|}{{\color[HTML]{9B9B9B} 2.19}} & \multicolumn{1}{c|}{39.8} & 10          \\
32-0.25x & \multicolumn{1}{c|}{{\color[HTML]{9B9B9B} 0.22}} & \multicolumn{1}{c|}{{\color[HTML]{9B9B9B} 0.44}} & \multicolumn{1}{c|}{13.4} & 23          &  & \multicolumn{1}{c|}{{\color[HTML]{9B9B9B} 0.51}} & \multicolumn{1}{c|}{{\color[HTML]{9B9B9B} 2.05}} & \multicolumn{1}{c|}{43.4} & 0           \\ \hline
\end{tabular}
\label{tbl:net_width}
\vspace{-0.4cm}
\end{table}

\begin{table}[htb!]
\renewcommand{\tabcolsep}{1.5pt}
\centering
\caption{Layerwise \% Sparsity by $\gamma$ for VGG-16 on CIFAR10 and 100. Also shown is the handcrafted sparse structure of \cite{li2017pruning}}
\resizebox{0.9\linewidth}{!}{
\begin{tabular}{cc|c|c|c|c|c|ccc}
                              &            & \multicolumn{4}{c|}{CIFAR 10}                                              &  & \multicolumn{3}{c}{CIFAR 100}               \\ \cline{3-10}
\multicolumn{1}{c|}{Conv}     &    \#Conv        & \multicolumn{3}{c|}{Adam, L2:1e-4}          &      {\color[HTML]{000199}Li et }                      &  & \multicolumn{3}{c}{Adam, L2:1e-4}          \\
\multicolumn{1}{c|}{Layer}    & Feat. & B: 40 & B: 80 & B: 160                     & {\color[HTML]{000199}al.\cite{li2017pruning} }         &  & \multicolumn{1}{c|}{B: 40}     & \multicolumn{1}{c|}{B: 80} & B: 160    \\ \hline
\multicolumn{1}{c|}{C1}       & 64         & 64     & 0     & {\color[HTML]{9a0000}0}                           & {\color[HTML]{009901}50}                          &  & \multicolumn{1}{c|}{49}         & 1    & \multicolumn{1}{|c}{58}     \\
\multicolumn{1}{c|}{C2}       & 64         & 18     & 0     & {\color[HTML]{9a0000}0}                           & {\color[HTML]{9a0000}0}                           &  & \multicolumn{1}{c|}{4}          & 0    & \multicolumn{1}{|c}{8}     \\
\multicolumn{1}{c|}{C3}       & 128        & 50     & 47    & {\color[HTML]{009901}51}                          & {\color[HTML]{9a0000}0}                           &  & \multicolumn{1}{c|}{29}         & 40    & \multicolumn{1}{|c}{54}    \\
\multicolumn{1}{c|}{C4}       & 128        & 12     & 5     & {\color[HTML]{9a0000}6}                           & {\color[HTML]{9a0000}0}                          &  & \multicolumn{1}{c|}{0}          & 0     & \multicolumn{1}{|c}{3}    \\
\multicolumn{1}{c|}{C5}       & 256        & 46     & 40    & {\color[HTML]{009901}36}                          & {\color[HTML]{9a0000}0}                           &  & \multicolumn{1}{c|}{10}         & 5     & \multicolumn{1}{|c}{27}    \\
\multicolumn{1}{c|}{C6}       & 256        & 71     & 66    & {\color[HTML]{009901}63}                          & {\color[HTML]{9a0000}0}                          &  & \multicolumn{1}{c|}{26}         & 5     & \multicolumn{1}{|c}{7}    \\
\multicolumn{1}{c|}{C7}       & 256        & 82     & 80    & {\color[HTML]{009901}79}                         & {\color[HTML]{9a0000}0}                          &  & \multicolumn{1}{c|}{44}         & 12    & \multicolumn{1}{|c}{0}    \\
\multicolumn{1}{c|}{C8}       & 512        & 95     & 96    & {\color[HTML]{009901}96}                          & {\color[HTML]{009901}50}                         &  & \multicolumn{1}{c|}{86}         & 74    & \multicolumn{1}{|c}{55}    \\
\multicolumn{1}{c|}{C9}       & 512        & 97     & 97    & {\color[HTML]{009901}97}                          & {\color[HTML]{009901}50}                          &  & \multicolumn{1}{c|}{95}         & 90    & \multicolumn{1}{|c}{94}    \\
\multicolumn{1}{c|}{C10}      & 512        & 97     & 97    & {\color[HTML]{009901}96}                         & {\color[HTML]{009901}50}                         &  & \multicolumn{1}{c|}{96}         & 93    &  \multicolumn{1}{|c}{93}  \\
\multicolumn{1}{c|}{C11}      & 512        & 98     & 98    & {\color[HTML]{009901}98}                          & {\color[HTML]{009901}50}                          &  & \multicolumn{1}{c|}{98}         & 97    &  \multicolumn{1}{|c}{96}    \\
\multicolumn{1}{c|}{C12}      & 512        & 99     & 99    & {\color[HTML]{009901}98}                          & {\color[HTML]{009901}50}                          &  & \multicolumn{1}{c|}{98}         & 98    &  \multicolumn{1}{|c}{99}    \\
\multicolumn{1}{c|}{C13}      & 512        & 99     & 99    & {\color[HTML]{009901}99}                          & {\color[HTML]{009901}50}                          &  & \multicolumn{1}{c|}{98}         & 98    &  \multicolumn{1}{|c}{96}    \\ \hline
\multicolumn{2}{c|}{\%Feat. Pruned}                & 86     & 84    & {\color[HTML]{009901}83}  & {\color[HTML]{009901}37}  &  & \multicolumn{1}{c|}{76}         & 69   &  \multicolumn{1}{|c}{69}     \\ \hline \hline
\multicolumn{2}{c|}{Test Err}             & 7.2    & 7.0   & \cellcolor[HTML]{9AFF99}6.5 & \cellcolor[HTML]{9AFF99}6.6 &  & \multicolumn{1}{c|}{29.2}       & 28.1 &   \multicolumn{1}{|c}{27.8}   
\end{tabular}
}

\label{tbl:vgg_cifar}
\vspace{-0.4cm}
\end{table}
\begin{table}[ht!]
\renewcommand{\tabcolsep}{2.5pt}
\centering
\caption{Sparsity by $\gamma$ on VGG-16, trained on TinyImageNet, and on ImageNet. Also shown are the pre- and post-pruning top-1/top-5 single crop validation errors. Pruning using $|\gamma| < 10^{-3}$ criteria.}
\label{tbl:vgg16_tinyimagenet}
\resizebox{0.95\linewidth}{!}{
\begin{tabular}{c|c|cc|c|c|}
             &        \# Conv               & \multicolumn{2}{c|}{Pre-pruning}    & \multicolumn{2}{c|}{Post-pruning} \\
TinyImageNet                &  Feat. Pruned & \multicolumn{1}{c|}{top1}  & top5  & top1          & top5        \\ \hline
L2: 1e-4, B: 20 & 3016 (71\%)           & \multicolumn{1}{c|}{45.1} & 21.4 & 45.1         &     21.4       \\
L2: 1e-4, B: 40 & 2571 (61\%)           & \multicolumn{1}{c|}{46.7} & 24.4 & 46.7         &     24.4       \\ \hline
 ImageNet    &       \multicolumn{5}{c|}{} \\ \hline
L2: 1e-4, B: 40 & 292           & \multicolumn{1}{c|}{29.93} & 10.41 & 29.91         &     10.41       \\
\end{tabular}
}
\label{tbl:vgg16_tinyimagenet}
\vspace{-0.4cm}
\end{table}
\begin{table}[htb!]
\renewcommand{\tabcolsep}{2.5pt}
\centering
\caption{Effect of different mini-batch sizes on sparsity (by $\gamma$) in VGG-11, trained on ImageNet. Same network structure employed as \cite{liu2017learning}. * indicates finetuning after pruning}
\label{tbl:vgg11_imagenet}
\resizebox{\linewidth}{!}{
\begin{tabular}{l|c|cc|c|c|}
                &       \# Conv               & \multicolumn{2}{c|}{Pre-pruning}    & \multicolumn{2}{c|}{Post-pruning} \\
                &  Feat. Pruned & \multicolumn{1}{c|}{top1}  & top5  & top1          & top5        \\ \hline
Adam, L2: 1e-4, B: 90 & {\color[HTML]{009901}71}                     & \multicolumn{1}{c|}{30.50} & 10.65 & \cellcolor[HTML]{9AFF99}30.47         &     10.64       \\
Adam, L2: 1e-4, B: 60 & {\color[HTML]{009901}140}                    & \multicolumn{1}{c|}{31.76} & 11.53 & 31.73         &     11.51        \\ \hline
{\color[HTML]{000199}Liu et al. \cite{liu2017learning} from \cite{liu2018rethinking}}      & {\color[HTML]{009901}85}                     & \multicolumn{1}{c|}{29.16} &       & 31.38*         &   -         
\end{tabular}
}
\label{tbl:vgg11_imagenet}
\vspace{-0.4cm}
\end{table}
\begin{table}[htb!]
\renewcommand{\tabcolsep}{1.5pt}
\centering
\caption{Sparsity by $\gamma$ on VGG-19, trained on CIFAR10/100. Also shown are the post-pruning test error. Compared with explicit sparsification approach of Liu et al. \cite{liu2017learning} }
\label{tbl:vgg19_cifar}
\resizebox{\linewidth}{!}{
\begin{tabular}{c|c|c|c|c|ccc}
                                          & \multicolumn{3}{c|}{CIFAR 10}                                              &  & \multicolumn{3}{c}{CIFAR 100}               \\ \cline{1-8}
\multicolumn{1}{c|}{}            & \multicolumn{2}{c|}{Adam, L2:1e-4}          &      {\color[HTML]{000199}Liu et}                       &  & \multicolumn{2}{c}{Adam, L2:1e-4} & {\color[HTML]{000199}Liu et  }       \\
\multicolumn{1}{c|}{}     & B: 64 & B: 512&      {\color[HTML]{000199}al.\cite{liu2017learning}}          &  & \multicolumn{1}{c|}{B: 64}     & \multicolumn{1}{c|}{B: 512} &  {\color[HTML]{000199}al.\cite{liu2017learning}}    \\ \hline
\multicolumn{1}{c|}{\%Feat. Pruned}                & 85     & {\color[HTML]{009901}81}    & {\color[HTML]{009901}70}    &  & \multicolumn{1}{c|}{75}         & {\color[HTML]{009901}62}   &  \multicolumn{1}{|c}{\color[HTML]{009901}50}     \\ \hline \hline
\multicolumn{1}{c|}{Test Err}             & 7.1    & 6.9   & \cellcolor[HTML]{9AFF99}6.3  &  & \multicolumn{1}{c|}{29.9}       & 28.8 &   \multicolumn{1}{|c}{\cellcolor[HTML]{9AFF99}26.7}   
\end{tabular}
}
\label{tbl:vgg19_cifar}
\vspace{-0.5cm}
\end{table}

\subsection{Comparison With Explicit Feature Sparsification / Pruning Approaches}
For VGG-16, we compare the network trained on CIFAR-10 with Adam using different mini-batch sizes against the handcrafted approach of Li et al.~\cite{li2017pruning}. Similar to tuning the explicit sparsification hyperparameter in~\cite{liu2017learning}, the mini-batch size can be varied to find the sparsest representation with an acceptable level of test performance. We see from Table~\ref{tbl:vgg_cifar} that when trained with a batch size of 160, 83\% of the features can be pruned away and leads to a better performance that the 37\% of the features pruned for ~\cite{li2017pruning}.
For VGG-11 on ImageNet (Table~\ref{tbl:vgg11_imagenet}), by simply varying the mini-batch size from 90 to 60, the number of convolutional features pruned goes from 71 to 140. This is in the same range as the number of features pruned by the explicit sparsification approach of ~\cite{li2017pruning}, and gives a comparable top-1 and top-5 validation error. 
For VGG-19 on CIFAR10 and CIFAR100 (Table~\ref{tbl:vgg19_cifar}), we see again that varying the mini-batch size controls the extent of sparsity. For the mini-batch sizes we considered, the extent of sparsity is much higher than that of \cite{liu2017learning}, with consequently slightly worse performance. The mini-batch size or other hyper-parameters can be tweaked to further tradeoff sparsity for accuracy, and reach a comparable sparsity-accuracy point as~\cite{liu2017learning}. 
\section{Discussion and Future Work}
Our findings relate to the anecdotally known and poorly understood `dying ReLU' phenomenon \cite{StanfordCS231n}, wherein some features in ReLU networks get cut off while training, leading to a reduced effective learning capacity of the network. 
Ameliorating it with Leaky ReLU \cite{maas2013rectifier} is ineffective because it does not address the root cause. \emph{BasicNet} with Leaky ReLU (negative slope of 0.01) on CIFAR-100 only marginally reduces the extent of sparsity in the case of Adam with L2: $10^{-4}$ (41\% feature sparsity vs. 47\% with ReLU). Reducing the learning rate of BN parameter $\gamma$ is much more effective (33\% sparsity). See Tables 2, 3 in the supplemental document.

Our work opens several avenues of future investigation. Understanding why features learned with Adam (and perhaps other adaptive methods) are more selective than with (m)SGD can further shed light on the practical differences between adaptive methods and SGD.
Also, our insights will lead practitioners to be more aware of the implicit tradeoffs between network capacity and generalization being made below the surface, while changing hyperparameters such as mini-batch size, which are seemingly unrelated to network capacity.
Also, we show that Adam with L2 regularization works out of the box for speeding up neural networks and is a strong baseline for future efforts towards filter-sparsification-for-speedup approaches.


\section{Conclusion}
We show through extensive experiments that the root cause for the emergence of filter level sparsity in CNNs is likely the disproportionate regularization (L2 or weight decay) of the parameters in comparison to the gradient from the primary objective. We identify how various factors influence the extent of sparsity by interacting in subtle ways with the regularization process. 
We show that adaptive gradient updates play a crucial role in the emergent sparsity (in contrast to SGD), and Adam not only shows a higher degree of sparsity but the extent of sparsity also has a strong dependence on the mini-batch size. We show that this is caused by the propensity of Adam to learn more selective features, and the added acceleration of L2 regularization interacting with the adaptive updates in low gradient regime.

Due to its targeting of selective features, the emergent sparsity can be used to trade off between network capacity, performance and generalization ability as per the task setting, and common hyperparameters such as mini-batch size allow direct control over it. We leverage this finegrained control and show that Adam with L2 regularization can be an attractive alternative to explicit network slimming approaches for speeding up test time performance of CNNs, without any tooling changes to the traditional neural network training pipeline supported by popular frameworks.

\clearpage
\begin{center}
\textbf{\large Supplementary Document:\\On Implicit Filter Level Sparsity\\In Convolutional Neural Networks}
\end{center}
\setcounter{equation}{0}
\setcounter{figure}{0}
\setcounter{table}{0}
\setcounter{section}{0}
In this supplemental document, we provide additional experiments that show how filter level sparsity manifests under different gradient descent flavours and regularization settings (Sec.~\ref{s:layerwisesparsity}), and that it even manifests with Leaky ReLU. We also show the emergence of feature selectivity in Adam in multiple layers, and discuss its implications on the extent of sparsity (Sec.~\ref{s:featureselectivity}). In Section~\ref{sec:other_hyp} we consider additional hyperparameters that influence the emergent sparsity. 
In Section~\ref{s:implementationdetails} we provide specifics for some of the experiments reported in the main document.

\begin{table*}[!bhp]
\renewcommand{\tabcolsep}{1.5pt}
\centering
\caption{Layerwise \% filters pruned from BasicNet trained on CIFAR10, based on the $|\gamma| <10^{-3}$ criteria. Also shown are pre-pruning and post-pruning test error, and the \% of \emph{convolutional} parameters pruned. C1-C7 indicate Convolution layer 1-7, and the numbers in parantheses indicate the total number of features per layer. Average of 3 runs. Also see Table 2 in the main document.}
\label{tbl:layerwise_sparsity_cifar10}
\resizebox{0.8\linewidth}{!}{
\begin{tabular}{lcccccccccccccc}
                    \multicolumn{2}{l}{\textbf{CIFAR10}}                                & \multicolumn{1}{l}{} & \multicolumn{1}{l}{}           & \multicolumn{1}{l|}{}          & \multicolumn{8}{c|}{\% Sparsity by $\gamma$ or \% Filters Pruned}                                                                                                                                                                & \multicolumn{1}{l|}{\% Param}                                          & \multicolumn{1}{l}{}         \\ \cline{6-13}
                       & \multicolumn{1}{l|}{}         &    Train                  & \multicolumn{1}{c|}{Test}          & \multicolumn{1}{c|}{Test}          & C1                   & C2                   & C3                   & C4                   & C5                   & C6                   & \multicolumn{1}{c|}{C7}    & \multicolumn{1}{c|}{Total}  & \multicolumn{1}{c|}{Pruned}                                            & Pruned                       \\
               & \multicolumn{1}{l|}{}         & Loss           & \multicolumn{1}{c|}{Loss} & \multicolumn{1}{c|}{Err}  & (64)                 & (128)                & (128)                & (256)                & (256)                & (512)                & \multicolumn{1}{c|}{(512)} & \multicolumn{1}{c|}{(1856)} & \multicolumn{1}{c|}{(4649664)}                                           & Test Err.                    \\ \hline
\multicolumn{1}{|c}{}                       & \multicolumn{1}{|c|}{L2: 1e-3} & 0.29                 & \multicolumn{1}{c|}{0.41}      & \multicolumn{1}{c|}{13.1}      & 59                   & 57                   & 42                   & 74                   & 76                   & 97                   & \multicolumn{1}{c|}{98}    & \multicolumn{1}{c|}{83}     & \multicolumn{1}{c|}{97}                        & 13.5 \\
\multicolumn{1}{|c}{}                       & \multicolumn{1}{|c|}{L2: 1e-4} & 0.06                 & \multicolumn{1}{c|}{0.43}      & \multicolumn{1}{c|}{10.5}      & 44                   & 22                   & 6                    & 45                   & 54                   & 96                   & \multicolumn{1}{c|}{95}    & \multicolumn{1}{c|}{70}     & \multicolumn{1}{c|}{{\color[HTML]{000000} 90}} & 10.5 \\ \cline{2-15} 
\multicolumn{1}{|c}{}                       & \multicolumn{1}{|c|}{WD: 2e-4} & 0.22                 & \multicolumn{1}{c|}{0.42}      & \multicolumn{1}{c|}{13.4}      & 57                   & 27                    & 9                    & 19                    & 46                   & 77                   & \multicolumn{1}{c|}{91}    & \multicolumn{1}{c|}{60}     & \multicolumn{1}{c|}{{\color[HTML]{000000} 83}} & 13.4 \\
\multicolumn{1}{|c}{\multirow{-4}{*}{\rotatebox[origin=c]{90}{Adam \B}}} & \multicolumn{1}{|c|}{WD: 1e-4} & 0.07                 & \multicolumn{1}{c|}{0.42}      & \multicolumn{1}{c|}{11.2}      & 45                    & 4                    & 0                    & 0                    & 14                    & 51                    & \multicolumn{1}{c|}{78}     & \multicolumn{1}{c|}{40}      & \multicolumn{1}{c|}{63}                         & 11.2 \\ \hline
\multicolumn{1}{|c}{}                       & \multicolumn{1}{|c|}{L2: 1e-3} & 0.62                 & \multicolumn{1}{c|}{0.64}      & \multicolumn{1}{c|}{21.8}      & 86                   & 61                   & 53                   & 46                   & 65                   & 4                    & \multicolumn{1}{c|}{0}     & \multicolumn{1}{c|}{27}     & \multicolumn{1}{c|}{38}                                                & 21.8                         \\
\multicolumn{1}{|c}{}                       & \multicolumn{1}{|c|}{L2: 5e-4} & 0.38                 & \multicolumn{1}{c|}{0.49}      & \multicolumn{1}{c|}{16.3}      & 68                    & 16                    & 9                    & 9                    & 24                    & 0                    & \multicolumn{1}{c|}{0}     & \multicolumn{1}{c|}{9}      & \multicolumn{1}{c|}{13}                         & 16.5 \\ \cline{2-15} 
\multicolumn{1}{|c}{}                       & \multicolumn{1}{|c|}{WD: 1e-3} & 0.61                 & \multicolumn{1}{c|}{0.63}      & \multicolumn{1}{c|}{21.6}      & 85                   & 60                   & 51                   & 46                   & 66                   & 4                   & \multicolumn{1}{c|}{0}    & \multicolumn{1}{c|}{27}     & \multicolumn{1}{c|}{{\color[HTML]{000000}38}} & 21.6 \\
\multicolumn{1}{|c}{\multirow{-4}{*}{\rotatebox[origin=c]{90}{SGD}}}  & \multicolumn{1}{|c|}{WD: 5e-4} & 0.38                 & \multicolumn{1}{c|}{0.46}      & \multicolumn{1}{c|}{15.8}      & 69                    & 19                    & 7                    & 7                    & 23                    & 0                    & \multicolumn{1}{c|}{0}     & \multicolumn{1}{c|}{8}      & \multicolumn{1}{c|}{13}                         & 16.1 \\ \hline
                                            & \multicolumn{1}{l}{}          & \multicolumn{1}{l}{} & \multicolumn{1}{l}{}           & \multicolumn{1}{l}{}           & \multicolumn{1}{l}{} & \multicolumn{1}{l}{} & \multicolumn{1}{l}{} & \multicolumn{1}{l}{} & \multicolumn{1}{l}{} & \multicolumn{1}{l}{} & \multicolumn{1}{l}{}       & \multicolumn{1}{l}{}        & \multicolumn{1}{l}{}                                                   & \multicolumn{1}{l}{}         \\
\end{tabular}
}
\vspace{-0.5cm}
\end{table*}
\renewcommand{\tabcolsep}{1.5pt}
\begin{table*}[]
\centering
\caption{Layerwise \% filters pruned from BasicNet trained on CIFAR100, based on the $|\gamma| <10^{-3}$ criteria. Also shown are pre-pruning and post-pruning test error. C1-C7 indicate Convolution layer 1-7, and the numbers in parantheses indicate the total number of features per layer. Average of 3 runs.}
\label{tbl:layerwise_sparsity_leaky}
\begin{tabular}{lccccccccccccc}
                                             \multicolumn{5}{l|}{\textbf{Adam vs AMSGrad (ReLU)}}                      & \multicolumn{8}{c|}{\% Sparsity by $\gamma$ or \% Filters Pruned}                                                                                                                                                                &                                            \multicolumn{1}{l}{}         \\ \cline{6-13}
                       & \multicolumn{1}{l|}{}         &    Train                  & \multicolumn{1}{c|}{Test}          & \multicolumn{1}{c|}{Test}          & \T C1                   & C2                   & C3                   & C4                   & C5                   & C6                   & \multicolumn{1}{c|}{C7}    & \multicolumn{1}{c|}{Total}                                               & Pruned                       \\
               & \multicolumn{1}{l|}{}         & Loss           & \multicolumn{1}{c|}{Loss} & \multicolumn{1}{c|}{Err}  & (64)                 & (128)                & (128)                & (256)                & (256)                & (512)                & \multicolumn{1}{c|}{(512)} & \multicolumn{1}{c|}{(1856)}                                            & Test Err.                    \\ \hline
\multicolumn{1}{|l}{}                       & \multicolumn{1}{|c|}{\T L2: 1e-3} & 1.06                 & \multicolumn{1}{c|}{1.41}      & \multicolumn{1}{c|}{39.0}      & 56                   & 47                   & 43                   & 68                   & 72                   & 91                   & \multicolumn{1}{c|}{85}    & \multicolumn{1}{c|}{76}                             & 39.3 \\
\multicolumn{1}{|c}{\multirow{-2}{*}{\rotatebox[origin=c]{90}{Adam}}}                       & \multicolumn{1}{|c|}{L2: 1e-4 \B} & 0.10                 & \multicolumn{1}{c|}{1.98}      & \multicolumn{1}{c|}{36.6}      & 41                   & 20                   & 9                    & 33                   & 34                   & 67                   & \multicolumn{1}{c|}{55}    & \multicolumn{1}{c|}{47}                            & 36.6 \\ \hline 
\multicolumn{1}{|l}{}                       & \multicolumn{1}{|c|}{\T L2: 1e-2} & 3.01                 & \multicolumn{1}{c|}{2.87}      & \multicolumn{1}{c|}{71.9}      & 79                   & 91                   & 91                   & 96                   & 96                   & 98                    & \multicolumn{1}{c|}{96}    & \multicolumn{1}{c|}{95}                             & 71.9 \\
\multicolumn{1}{|l}{}                       & \multicolumn{1}{|c|}{\T L2: 1e-4} & 0.04                 & \multicolumn{1}{c|}{1.90}      & \multicolumn{1}{c|}{35.6}      & 0                    & 0                    & 0                    & 0                    & 1                    & 25                    & \multicolumn{1}{c|}{23}     & \multicolumn{1}{c|}{13}                               & 35.6 \\  
\multicolumn{1}{|c}{\multirow{-3}{*}{\rotatebox[origin=c]{90}{AMSGrad}}}  & \multicolumn{1}{|c|}{\T L2: 1e-6\B} & 0.01                 & \multicolumn{1}{c|}{3.23}      & \multicolumn{1}{c|}{40.2}      & 0                    & 0                    & 0                   & 0                    & 0                    & 0                    & \multicolumn{1}{c|}{0}     & \multicolumn{1}{c|}{0}                               & 40.2 \\ \hline
\end{tabular}
\begin{tabular}{ccccccccccccc}
\\
                                             \multicolumn{4}{l|}{\textbf{Adam With Leaky ReLU}}             & \multicolumn{8}{c|}{\% Sparsity by $\gamma$ or \% Filters Pruned}                                                                                                                                                                &                                            \multicolumn{1}{l}{}         \\ \cline{5-12}
                        \multicolumn{1}{l|}{}         &    Train                  & \multicolumn{1}{c|}{Test}          & \multicolumn{1}{c|}{Test}          & \T C1                   & C2                   & C3                   & C4                   & C5                   & C6                   & \multicolumn{1}{c|}{C7}    & \multicolumn{1}{c|}{Total}                                               & Pruned                       \\
                \multicolumn{1}{l|}{NegSlope=0.01}         & Loss           & \multicolumn{1}{c|}{Loss} & \multicolumn{1}{c|}{Err}  & (64)                 & (128)                & (128)                & (256)                & (256)                & (512)                & \multicolumn{1}{c|}{(512)} & \multicolumn{1}{c|}{(1856)}                                            & Test Err.                    \\ \hline
 \multicolumn{1}{c|}{L2: 1e-3} & 1.07                 & \multicolumn{1}{c|}{1.41}      & \multicolumn{1}{c|}{39.1}      & 49                   & 40                   & 39                   & 62                   & 61                   & 81                   & \multicolumn{1}{c|}{85}    & \multicolumn{1}{c|}{70}                             & 39.4 \\
 \multicolumn{1}{c|}{L2: 1e-4} & 0.10                 & \multicolumn{1}{c|}{1.99}      & \multicolumn{1}{c|}{36.8}      & 33                   & 20                   & 9                    & 31                   & 29                   & 55                   & \multicolumn{1}{c|}{53}    & \multicolumn{1}{c|}{41}                            & 36.8 \\ \hline 
 \multicolumn{13}{l}{NegSlope=0.1} \\ \hline
 \multicolumn{1}{c|}{L2: 1e-4} & 0.14                 & \multicolumn{1}{c|}{2.01}      & \multicolumn{1}{c|}{37.2}      & 38                   & 30                   & 21                    & 34                   & 31                   & 55                   & \multicolumn{1}{c|}{52}    & \multicolumn{1}{c|}{43}                            & 37.3 \\ \hline 
\end{tabular}
\vspace{-0.3cm}
\end{table*}

\section{Layer-wise Sparsity in \emph{BasicNet}}
\label{s:layerwisesparsity}
In Section 2.3 and Table 2 in the main paper, we demonstrated that for BasicNet on CIFAR-100, Adam shows feature sparsity in both early layers and later layers, while SGD only shows sparsity in the early layers. We establish in the main paper that Adam learns selective features in the later layers which contribute to this additional sparsity. 
In Table~\ref{tbl:layerwise_sparsity_cifar10} we show similar trends in layer-wise sparsity also emerge when trained on CIFAR-10. 

\parahead{Sparsity with AMSGrad} In Table \ref{tbl:layerwise_sparsity_leaky} we compare the extent of sparsity of Adam with AMSGrad~\cite{reddi2018convergence}. Given that AMSGrad tracks the long term history of squared gradients, we expect the effect of L2 regularization in the low gradient regime to be dampened, and for it to lead to less sparsity. For BasicNet, on CIFAR-100, with L2 regularization of $10^{-4}$, AMSGrad only shows sparsity in the later layers, and overall only 13\% of features are inactive. For a comparable test error for Adam, 47\% of the features are inactive. In Table~\ref{tbl:extra_optim_supp} we show the feature sparsity by activation and by $\gamma$ for BasicNet with AMSGrad, Adamax and RMSProp, trained for CIFAR-10/100.

\parahead{Sparsity with Leaky ReLU} Leaky ReLU is anecdotally~\cite{StanfordCS231n} believed to address the `dying ReLU' problem by preventing features from being inactivated. The cause of feature level sparsity is believed to be the accidental inactivation of features, which gradients from Leaky ReLU can help revive. We have however shown there are systemic processes underlying the emergence of feature level sparsity, and those would continue to persist even with Leaky ReLU. Though our original definition of feature selectivity does not apply here, it can be modified to make a distinction between data points which produce positive activations for a feature vs. the data points that produce a negative activation. 
For typical values of the negative slope (0.01 or 0.1) of Leaky ReLU, the more selective features (as per the updated definition) would continue to see lower gradients than the less selective features, and would consequently see relatively higher effect of regularization. For BasicNet trained on CIFAR-100 with Adam, in Table \ref{tbl:layerwise_sparsity_leaky} we see that using Leaky ReLU has a minor overall impact on the emergent sparsity. See Section~\ref{sec:other_hyp} for more effective ways of reducing filter level sparsity in ReLU networks.

\section{On Feature Selectivity in Adam}
\label{s:featureselectivity}
\begin{figure*}
  \centering
  \includegraphics[width=1.0\linewidth]{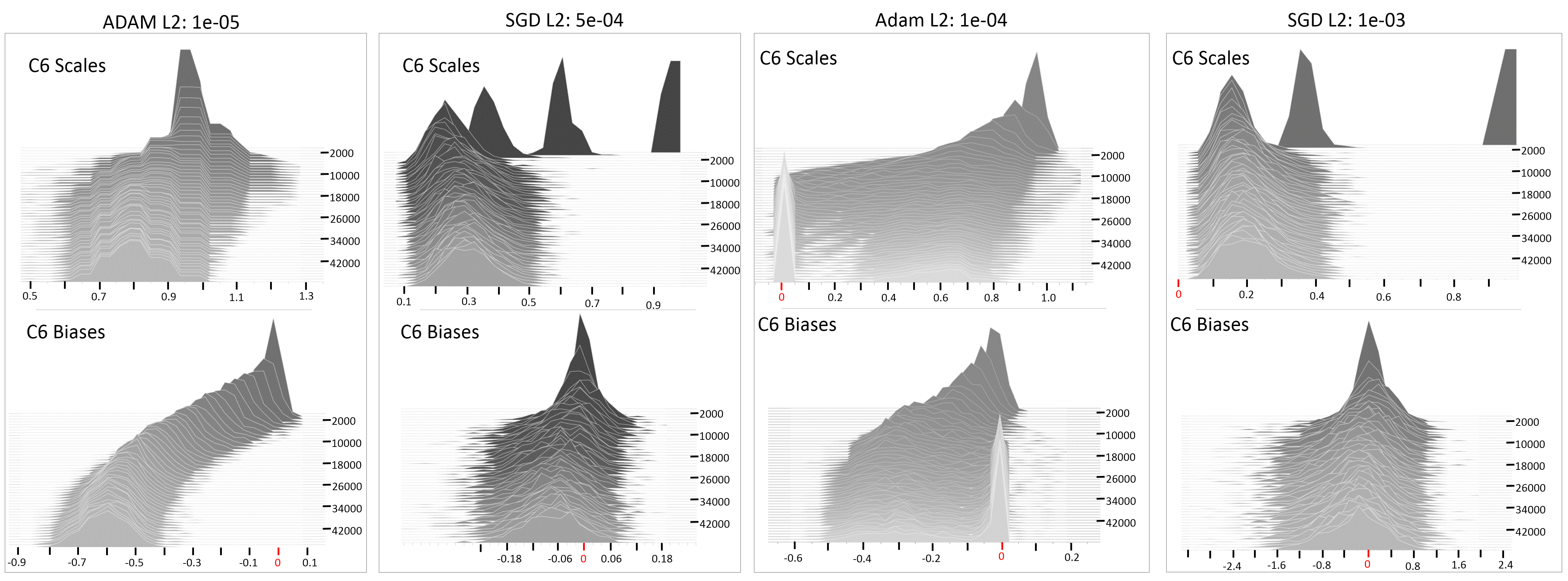}
  \includegraphics[width=1.0\linewidth]{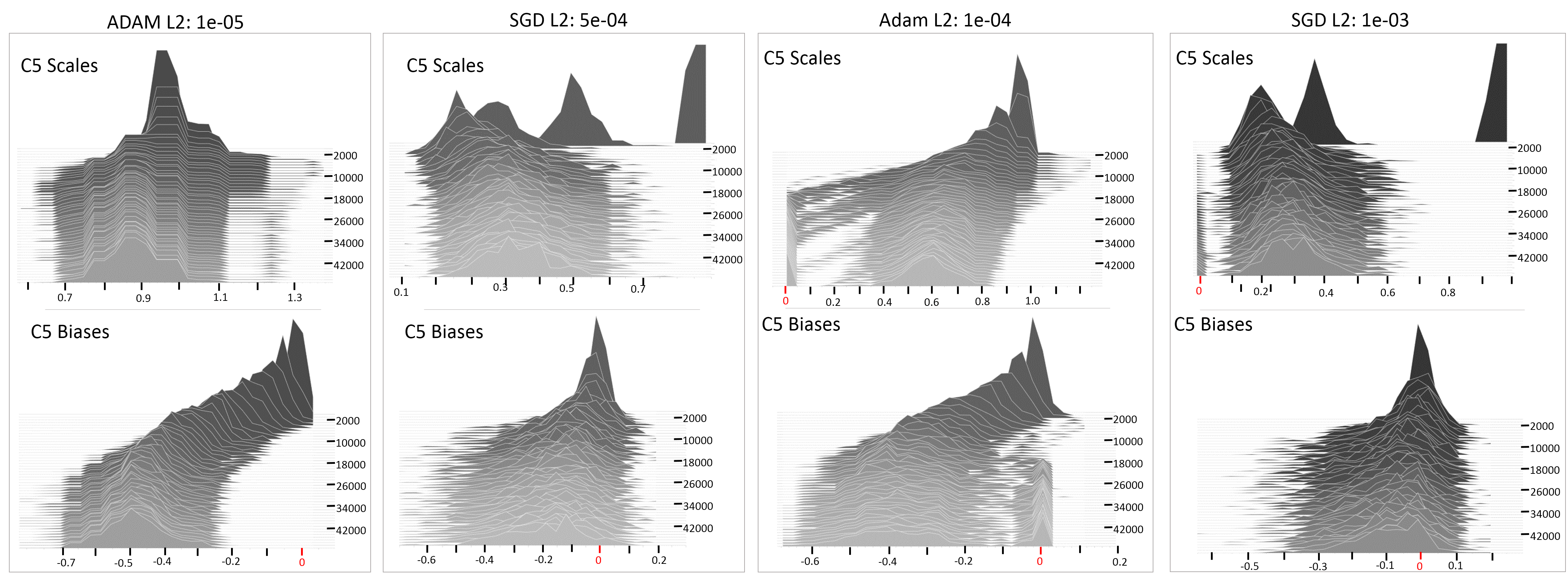}
  \caption{\textbf{Emergence of Feature Selectivity with Adam (Layer C6 and C5)} The evolution of the learned scales ($\gamma$, top row) and biases ($\beta$, bottom row) 
for layer C6 (top) and C5 (bottom) of \emph{BasicNet} for Adam and SGD as training progresses, in both low and high L2 regularization regimes. Adam has distinctly negative biases, while SGD sees both positive and negative biases. For positive scale values, as seen for both Adam and SGD, this translates to greater feature selectivity in the case of Adam, which translates to a higher degree of sparsification when stronger regularization is used.}
  \label{fig:bias_scale_distribution_full_c6}
\end{figure*}

In Figure~\ref{fig:bias_scale_distribution_full_c6}, we show the the distribution of the scales ($\gamma$) and biases ($\beta$) of layers C6 and C5 of \emph{BasicNet}, trained on CIFAR-100. We consider SGD and Adam, each with a low and high regularization value. For both C6 and C5, Adam learns exclusively negative biases and positive scales, which results in features having a higher degree of selectivity (i.e, activating for only small subsets of the training corpus). In case of SGD, a subset of features learns positive biases, indicating more universal (less selective) features. 

\begin{figure*}
\centering
  \includegraphics[width=1.0\linewidth]{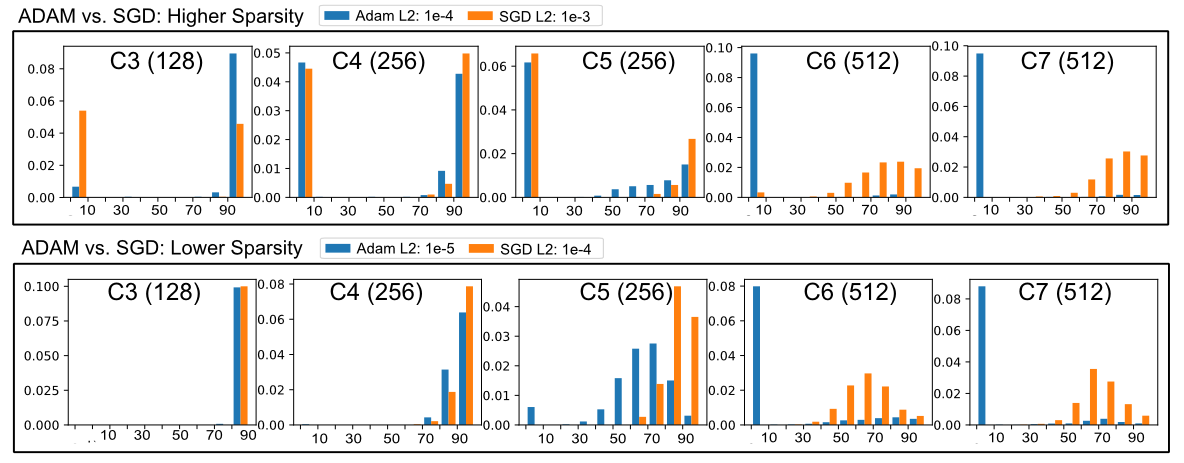}
  \caption{\textbf{Layer-wise Feature Selectivity} Feature universality for CIFAR 10, with Adam and SGD. X-axis shows the universality and Y-axis ($\times 10$) shows the fraction of features with that level of universality. For later layers, Adam tends to learn less universal features than SGD, which get pruned by the regularizer. Please be mindful of the differences in Y-axis scales between plots. Figure 3 in the main document shows similar plots for CIFAR100.}
  \label{fig:cifar_10_feat_selectivity}
\end{figure*}

Figure~\ref{fig:cifar_10_feat_selectivity} shows feature selectivity also emerges in the later layers when trained on CIFAR-10, in agreement with the results presented for CIFAR-100 in Fig.~3 of the main paper.

\begin{figure*}
  \includegraphics[width=1.0\linewidth]{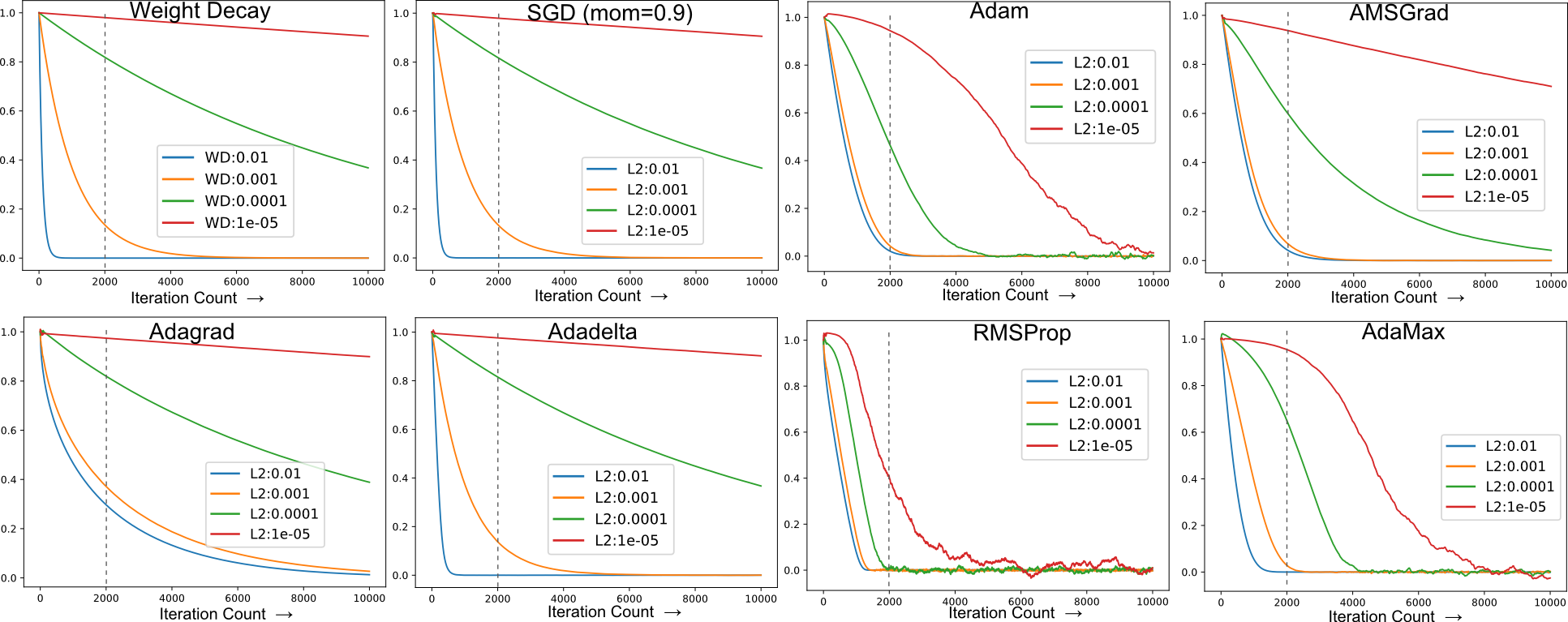}
  \caption{The action of regularization on a scalar value for a range of regularization values in the presence of simulated low gradients drawn from a mean=0, std=$10^{-5}$ normal distribution. The gradients for the first 100 iterations are drawn from a mean=0, std=$10^{-3}$ normal distribution to emulate a transition into low gradient regime rather than directly starting in the low gradient regime. The scalar is initialized with a value of 1. The learning rates are as follows: SGD(momentum=0.9,lr=0.1), ADAM(1e-3), AMSGrad(1e-3), Adagrad(1e-2), Adadelta(1.0), RMSProp(1e-3), AdaMax(2e-3). The action of the regularizer in low gradient regime is only one of the factors influencing sparsity. Different gradient descent flavours promote different levels of feature selectivity, which dictates the fraction of features that fall in the low gradient regime. Further, the optimizer and the mini-batch size affect together affect the duration different features spend in low gradient regime.}
  \label{fig:wd_l2_decay_full}
  \vspace{-0.3cm}
\end{figure*}

Higher feature selectivity leads to parameters spending more iterations in a low gradient regime.
In Figure~\ref{fig:wd_l2_decay_full}, we show the effect of the coupling of L2 regularization with the update step of various adaptive gradient descent approaches in a low gradient regime. Adaptive gradient approaches exhibit strong regularization in low gradient regime even with low regularization values. This disproportionate action of the regularizer, combined with the propensity of certain adaptive gradient methods for learning selective features, results in a higher degree of feature level sparsity with adaptive approaches than vanilla SGD, or when using weight decay.

\begin{figure*}
  \includegraphics[width=1.0\linewidth]{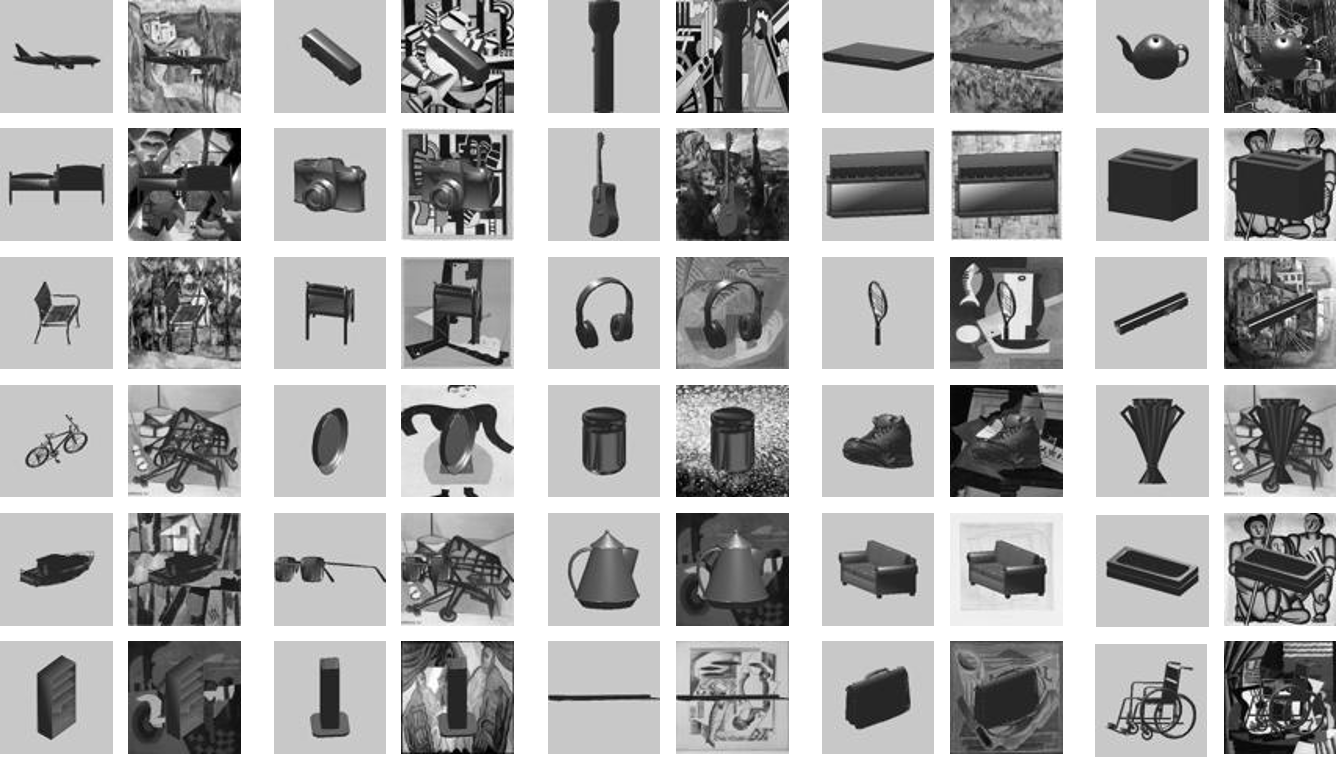}
  \caption{Unaugmented and augmented renderings of the subset of 30 classes from ObjectNet3D~\cite{xiang2016objectnet3d} employed to gauge the effect of task difficulty on implicit filter sparsity. The rendered images are 64x64 and obtained by randomly sampling (uniformly) the azimuth angle between -180 and 180 degrees, and the elevation between -15 and +45 degrees. The renderings are identical between the augmented and the unaugmented set and only differ in the background. The background images are grayscale versions of the Cubism subset from PeopleArt~\cite{wen2016learning} dataset.}
  \label{fig:objectnet3d}
  \vspace{-0.3cm}
\end{figure*}

\section{Effect of Other Hyperparameters on Sparsity}
\label{sec:other_hyp}
Having shown in the main paper and in Sec.~\ref{s:featureselectivity} that feature selectivity results directly from negative bias ($\beta$) values when the scale values ($\gamma$) are positive, we investigate the effect of $\beta$ initialization value on the resulting sparsity. As shown in Table~\ref{tbl:layerwise_sparsity_cifar100_scale_bias} for BasicNet trained with Adam on CIFAR 100, a slightly negative initialization value of $-$0.1 does not affect the level of sparsity. However, a positive initialization value of 1.0 results in higher sparsity. This shows that attempting to address the emergent sparsity by changing the initialization of $\beta$ may be counter productive.

We also investigate the effect of scaling down the learning rate of $\gamma$ and $\beta$ compared to that for the rest of the network (Table~\ref{tbl:layerwise_sparsity_cifar100_scale_bias}). Scaling down the learning rate of $\gamma$s by a factor of 10 results in a significant reduction of sparsity. This can likely be attributed to the decrease in effect of the L2 regularizer in the low gradient regime because it is directly scaled by the learning rate. This shows that tuning the learning of $\gamma$ can be more effective than Leaky ReLU at controlling the emergent sparsity. On the other hand, scaling down the learning rate of $\beta$s by a factor of 10 results in a slight increase in the extent of sparsity.
\renewcommand{\tabcolsep}{1.5pt}
\begin{table*}[]
\centering
\caption{Layerwise \% filters pruned from BasicNet trained on CIFAR100, based on the $|\gamma| <10^{-3}$ criteria. Also shown are pre-pruning and post-pruning test error. C1-C7 indicate Convolution layer 1-7, and the numbers in parantheses indicate the total number of features per layer. We analyse the effect of different initializations of $\beta$s, as well as the effect of different relative learning rates for $\gamma$s and $\beta$s, when trained with Adam with L2 regularization of $10^{-4}$. Average of 3 runs.}
\label{tbl:layerwise_sparsity_cifar100_scale_bias}
\begin{tabular}{ccccccccccccc}
\\
                                             \multicolumn{4}{l|}{\textbf{}}             & \multicolumn{8}{c|}{\% Sparsity by $\gamma$ or \% Filters Pruned}                                                                                                                                                                &                                            \multicolumn{1}{l}{}         \\ \cline{5-12}
                        \multicolumn{1}{l|}{}         &    Train                  & \multicolumn{1}{c|}{Test}          & \multicolumn{1}{c|}{Test}          & \T C1                   & C2                   & C3                   & C4                   & C5                   & C6                   & \multicolumn{1}{c|}{C7}    & \multicolumn{1}{c|}{Total}                                               & Pruned                       \\
                \multicolumn{1}{l|}{}         & Loss           & \multicolumn{1}{c|}{Loss} & \multicolumn{1}{c|}{Err}  & (64)                 & (128)                & (128)                & (256)                & (256)                & (512)                & \multicolumn{1}{c|}{(512)} & \multicolumn{1}{c|}{(1856)}                                            & Test Err.                    \\ \hline
 \multicolumn{1}{c|}{Baseline ($\gamma_{init}$=1, $\beta_{init}$=0)} & 0.10                 & \multicolumn{1}{c|}{1.98}      & \multicolumn{1}{c|}{36.6}      & 41                   & 20                   & 9                    & 33                   & 34                   & 67                   & \multicolumn{1}{c|}{55}    & \multicolumn{1}{c|}{46}                            & \T 36.6 \\ \hline 
 \multicolumn{1}{c|}{$\gamma_{init}$=1, $\beta_{init}$=$-$0.1} & 0.10                 & \multicolumn{1}{c|}{1.98}      & \multicolumn{1}{c|}{37.2}      & 44                   & 20                   & 10                    & 34                   & 32                   & 68                   & \multicolumn{1}{c|}{54}    & \multicolumn{1}{c|}{46}                            & \T 36.5 \\ \hline 
 \multicolumn{1}{c|}{$\gamma_{init}$=1, $\beta_{init}$=1.0} & 0.14                 & \multicolumn{1}{c|}{2.04}      & \multicolumn{1}{c|}{38.4}      & 47                   & 29                   & 25                    & 36                   & 46                   & 69                   & \multicolumn{1}{c|}{61}    & \multicolumn{1}{c|}{53}                            & \T 38.4 \\ \hline 
 \multicolumn{13}{l}{Different Learning Rate Scaling for $\beta$ and $\gamma$} \\ \hline
 \multicolumn{1}{c|}{LR scale for $\gamma$: 0.1} & 0.08                 & \multicolumn{1}{c|}{1.90}      & \multicolumn{1}{c|}{35.0}      & 16                   & 6                   & 1                    & 13                   & 20                   & 52                   & \multicolumn{1}{c|}{49}    & \multicolumn{1}{c|}{33}                            & \T 35.0 \\ \hline 
 \multicolumn{1}{c|}{LR scale for $\beta$: 0.1} & 0.12                 & \multicolumn{1}{c|}{1.98}      & \multicolumn{1}{c|}{37.1}      & 42                   & 26                   & 21                    & 41                   & 48                  & 70                   & \multicolumn{1}{c|}{55}    & \multicolumn{1}{c|}{51}                            & \T 37.1 \\ \hline 
\end{tabular}
\vspace{-0.3cm}
\end{table*}
\section{Experimental Details}
\label{s:implementationdetails}

For all experiments, the learned BatchNorm scales ($\gamma$) are initialized with a value of 1, and the biases ($\beta$) with a value of 0.
The reported numbers for all experiments on CIFAR10/100 are averaged over 3 runs. Those on TinyImageNet are averaged over 2 runs, and for ImageNet the results are from 1 run.
On CIFAR10/100, VGG-16 follows the same learning rate schedule as \emph{BasicNet}, as detailed in Section 2.1 in the main paper.

For experiments on ObjectNet3D~\cite{xiang2016objectnet3d} renderings, we use objects from the following 30 classes: aeroplane, bed, bench, bicycle, boat, bookshelf, bus, camera, chair, clock, eyeglasses, fan, flashlight, guitar, headphone, jar, kettle, keyboard, laptop, piano, racket, shoe, sofa, suitcase, teapot, toaster, train, trophy, tub, and wheelchair. The objects are rendered to 64x64 pixel images by randomly sampling (uniformly) the azimuth angle between -180 and 180 degrees, and the elevation between -15 and +45 degrees. The renderings are identical between the cluttered and the plain set, with the backgrounds for the cluttered set taken from the Cubism subset from PeopleArt~\cite{wen2016learning} dataset. See Figure~\ref{fig:objectnet3d}. The network structure and training is similar to that for CIFAR10/100, and a batch size of 40 is used.

On TinyImageNet, both VGG-16 and BasicNet follow similar schemes. Using a mini-batch size of 40, the gradient descent method specific base learning rate is used for 250 epochs, and scaled down by 10 for an additional 75 epochs and further scaled down by 10 for an additional 75 epochs, totaling 400 epochs. When the mini-batch size is adjusted, the number of epochs are appropriately adjusted to ensure the same number of iterations.

On ImageNet, the base learning rate for Adam is 1e-4. For \emph{BasicNet}, with a mini-batch size of 64, the base learning rate is used for 15 epochs, scaled down by a factor of 10 for another 15 epochs, and further scaled down by a factor of 10 for 10 additional epochs, totaling 40 epochs. The epochs are adjusted with a changing mini-batch size. For VGG-11, with a mini-batch size of 60, the total epochs are 60, with learning rate transitions at epoch 30 and epoch 50. 
For VGG-16, mini-batch size of 40, the total number of epochs are 50, with learning rate transitions at epoch 20 and 40.

\renewcommand{\tabcolsep}{1.5pt}
\begin{table}[t]
\centering
\caption{Convolutional filter sparsity in \emph{BasicNet} trained on CIFAR10/100 for Adamax, AMSGrad and RMSProp with L2 regularization. Shown are the \% of non-useful / inactive convolution filters, as measured by activation over training corpus (max act. $< 10^{-12}$) and by the learned BatchNorm scale ($|\gamma| < 10^{-03}$), averaged over 3 runs. 
See Table 1 in main paper for other combinations of regularization and gradient descent methods.}
\label{tbl:extra_optim_supp}
\resizebox{0.82\linewidth}{!}{
\begin{tabular}{c|c|c|c|c|c|c|c|c|}
\multicolumn{9}{c}{}\\
\cline{2-9}
                              &        & \multicolumn{3}{c}{CIFAR10}               &  &   \multicolumn{3}{c|}{CIFAR100}                      \\ \cline{3-9}
                              &        & \multicolumn{2}{c|}{\% Sparsity}    & Test           &  &   \multicolumn{2}{c|}{\% Sparsity} & Test                     \\
                              & \textbf{L2}       & by Act & by $\gamma$   & Error         &   & by Act & by $\gamma$   & Error              \\ \hline
                              \\
                               & 1e-02  & 93                          & 93                       & 20.9                                                                                  &  & 95                          & 95                       & 71.9                                                                                  \\
                               & 1e-04 & 51                           & 47                        & 9.9                                                                                  &  & 20                           & 13                        & 35.6                                                                                  \\
\multirow{-4}{*}{{\rotatebox[origin=c]{90}{AMSGrad}}}         & 1e-06 &  0     &  0  & 11.2                                                          &  & {0 }                           & {0}                         & 40.2                                                                                  \\ \hline 
                              \\
                               & 1e-02  & 75                          & 90                       & 16.4                                                                                  &  & 74                          & 87                      & 51.8                                                                                  \\
                               & 1e-04 & 49                           & 50                        & 10.1                                                                                  &  & 10                           & 10                        & 39.3                                                                                  \\
\multirow{-4}{*}{{\rotatebox[origin=c]{90}{Adamax}}}         & 1e-06 &  4     &  4  & 11.3                                                          &  & {0 }                           & {0}                         & 39.8                                                                                  \\ \hline 
                              \\
                               & 1e-02  & 95                          & 95                       & 26.9                                                                                  &  & 97                          & 97                       & 78.6                                                                                  \\
                               & 1e-04 & 72                           & 72                        & 10.4                                                                                  &  & 48                           & 48                        & 36.3                                                                                  \\
\multirow{-4}{*}{{\rotatebox[origin=c]{90}{RMSProp}}}         & 1e-06 &  29     &  29  & 10.9                                                          &  & {0 }                           & {0}                         & 40.6                                                                                  \\ \hline 
\end{tabular}
}
\end{table}

\vspace{10cm}
{\small
\bibliographystyle{ieee}
\bibliography{article}
}

\end{document}